\documentclass{article}

\usepackage[preprint]{neurips_2020}

\usepackage{natbib}

\usepackage[utf8]{inputenc} % allow utf-8 input
\usepackage[T1]{fontenc}    % use 8-bit T1 fonts
\usepackage{hyperref}       % hyperlinks
\usepackage{url}            % simple URL typesetting
\usepackage{booktabs}       % professional-quality tables
\usepackage{amsfonts}       % blackboard math symbols
\usepackage{nicefrac}       % compact symbols for 1/2, etc.
\usepackage{microtype}      % microtypography
\usepackage{amssymb}
\usepackage{xcolor}
\usepackage{graphicx}

\usepackage{amsthm}
\usepackage{amsmath}
\usepackage{algorithm}

\title{Buffer Zone based Defense against Adversarial Examples in Image Classification}

\author{Kaleel Mahmood$^{1*}$, Phuong Ha Nguyen$^{1*}$, Lam M. Nguyen$^{2}$, Thanh Nguyen$^{3}$, Marten van Dijk$^{1,4}$ \\
$^{1}$ Department of Electrical and Computer Engineering, University of Connecticut, USA\\
$^{2}$ IBM Research, Thomas J. Watson Research Center, Yorktown Heights, USA\\
$^{3}$ Iowa State University, USA \\
$^{4}$ CWI Amsterdam, The Netherlands\\
$^{*}$ Equally contributed \\
\texttt{phuongha.ntu@gmail.com}, \texttt{\{kaleel.mahmood, marten.van$\_$dijk\}@uconn.edu},  \\
\texttt{LamNguyen.MLTD@ibm.com}, \texttt{thanhng@iastate.edu}
}

\begin{document}

\maketitle

\begin{abstract}
We propose a novel defense against all existing gradient based adversarial attacks on deep neural networks for image classification problems. Our defense is based on a combination of deep neural networks and simple image transformations. While straightforward in implementation, this defense yields a unique security property which we term buffer zones. We argue that our defense based on buffer zones offers significant improvements over state-of-the-art defenses. We are able to achieve this improvement even when the adversary has access to the {\em entire} original training data set and unlimited query access to the defense. We verify our claim through experimentation using Fashion-MNIST and CIFAR-10: We demonstrate $<11\%$ attack success rate -- significantly lower than what other well-known state-of-the-art defenses offer -- at only a price of a $11-18\%$ drop in clean accuracy. By using a new intuitive metric, we explain why this trade-off offers a significant improvement over prior work.
\end{abstract}

\section{Introduction}\label{sec:intro}

There are many applications based on Convolution Neural Networks (CNNs) such as image classification \cite{KrizhevskySutskeverHinton2012,SimonyanZisserman2015}, object detection \cite{Girshick2015,RenHeGirshickEtAl2015}, semantic segmentation \cite{ShelhamerLongDarrell2017} and visual concept discovery \cite{WangZhangXieEtAl2017}. However, it is well-known that CNNs are highly susceptible to small perturbations $\eta$ which are added to \textit{benign} input images $x$. As shown in \cite{SzegedyZarembaSutskeverEtAl2013,GoodfellowShlensSzegedy2014}, by adding {\em visually imperceptible} perturbations to the original image, adversarial examples $x'$ can be created, i.e., $x'=x+\eta$. These adversarial examples are misclassified by the CNN with high confidence. Hence, making CNNs secure against this type of attack is a significantly important task. 
In the literature, adversarial machine learning attacks can be categorized as either white-box or black-box. This categorization depends on how much information about the classifier is necessary to run the attack. Due to the challenging nature of designing defenses against white-box attacks~\cite{tramer2020adaptive,AthalyeCarliniWagner2018,carlini2017adversarial}, the classifier parameters are kept/assumed secret in this paper. This disallows white-box attacks and so we focus exclusively on black-box adversaries. This setup is common, for example online machine learning services by default only allow black-box access to their models and do not publish their model parameters~\cite{PapernotMcDanielGoodfellowEtAl2017}.
\noindent

\begin{figure}%[!h]
\centering
\includegraphics[scale=0.25]{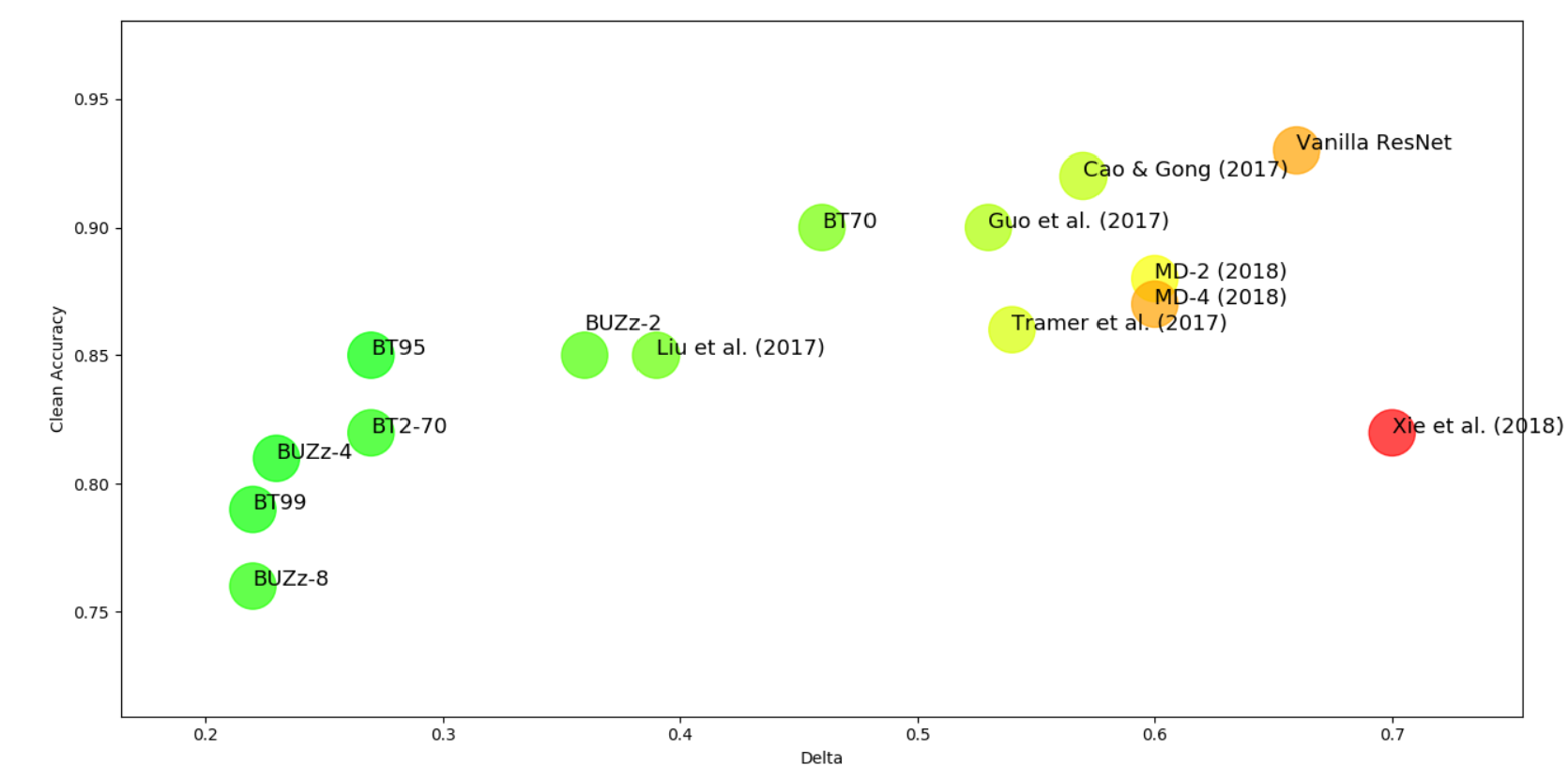}
\caption[]{CIFAR-10 defenses with clean accuracy $p_d$ (Clean Accuracy) and $\delta$ value (Delta) for untargeted black-box attacks. Acronyms for the various defenses are given in Section 5.}
 \label{fig:key_idea}
%\vspace{2mm}
\noindent
\end{figure}
\noindent

\begin{figure*}%[!h]
\centering
\includegraphics[scale=0.50]{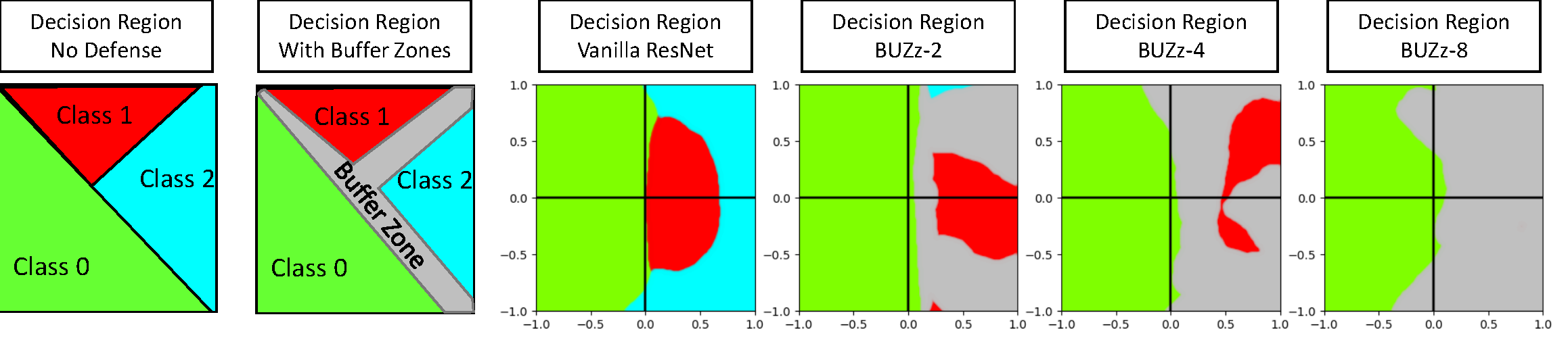}
\caption[]{Decision regions with and without buffer zones $\bot$.}
 \label{fig:fish}
%\vspace{2mm}
\noindent
\end{figure*}
\noindent

{\bf Defense performance metrics.}
Many defenses have been proposed in the literature to counter adversarial machine learning. Some popular recent defenses include~\cite{papernot2016distillation,KurakinGoodfellowBengio2016,TramerKurakinPapernotEtAl2017,CaoGong2017,MetzenGeneweinFischerEtAl2017,FeinmanCurtinShintreEtAl2017,XieWangZhangEtAl2017,meng2017magnet,GuoRanaCisseEtAl2017,srisakaokul2018muldef}. To properly analyze a defense two important factors must be considered. First, what is the attack success rate on the defense? Second, what kind of drop in clean accuracy is required to achieve the defense? To properly evaluate adversarial defenses, we explain in section \ref{sec:metric} a new metric $\delta$ which encapsulates both pieces of information: $\delta$ is defined as $\delta= \gamma+(p-\gamma)\alpha$, 
%\begin{align}
%    \label{eq_important}
%    \delta = p - (p-\gamma)(1-\alpha) = \gamma + (p-\gamma)\alpha. 
%\end{align}
where the original clean accuracy (of the vanilla scheme without defense) is denoted as $p$, the drop in clean accuracy caused by implementing the defense is denoted as $\gamma$ and the 
attack success rate is denoted as $\alpha$.
%new accuracy of the defense is $p_d $. 
%Essentially, a small $\delta$ is representative of a defense with a small attack success rate, as well as a small penalty in clean accuracy. 
The new clean accuracy of the defense (without attack) is $p_d=p-\gamma$. In section \ref{sec:metric} we show that the accuracy of the defense in the presence of adversaries is equal to $p-\delta$.
In Figure~\ref{fig:key_idea} we show  the $\delta$ vs clean accuracy $p_d$ for various ResNet defenses against untargeted black-box attacks for the CIFAR10 data set in the literature as well as our own proposed BUZz defenses. 

It is important to note that our defense not only achieves the lowest $\delta$, but also the various BUZz configurations offer different possible trade-offs between a higher clean prediction accuracy $p_d$ or a smaller $\delta$. %\textcolor{red}{Ha: 
An important lesson from~\cite{tramer2020adaptive,AthalyeCarliniWagner2018,carlini2017adversarial} is that it seems not possible to gain security (a small $\delta$ implied by a high defender success rate $1-\alpha$) with high clean accuracy ($p_d$), in other words, there must be a trade-off.
BUZz is the first method which can allow the user to tune security andclean prediction accuracy.
%} 

\textbf{Buffer zone based defense.}
Our defense is based on a novel concept which we term “buffer zones”. Buffer zones are regions in between classes with a special label. If an input image falls into this buffer zone region, the classifier returns a null label (i.e. the image is recognized as adversarial). In order for an adversary to create an image that defeats the defense, they must add a large perturbation to the input image. However, as mentioned in~\cite{papernot2016distillation} an attack is only considered successful if the adversarial noise $\eta$ has small magnitude, say $\|\eta\|\leq \epsilon$, which cannot be recognized by human beings.
The basic concept of buffer zones is illustrated in Figure~\ref{fig:fish}. 
%(see Supplemental Material \label{bufferzones} for details). 
In Figure~\ref{fig:fish} we show a hypothetical decision region without buffer zones and with buffer zones. We also show decision regions for a single sample from the CIFAR-10 testing set on a vanilla (no defense) ResNet56, as well as various ResNet BUZz defense configurations. The basic concept is the larger the buffer zone (the region in gray), the larger the perturbation required to change the input image from its original class label (in green) to another class label (in blue or red). %\textcolor{red}{Ha: 
Figure~\ref{fig:fish} is the proof-of-concept of the existence of buffer zones and it implies that any existing attack (i.e., white-box and black box attacks) should create sufficient large noises to cross the buffer zones. This fact allows us to achieve our security goal.
%}
A detailed description for how to generate the graphs in Figure~\ref{fig:fish} is in Supplemental Material \ref{bufferzones}. We explain how to create buffer zone based defenses later in the paper.  

{\bf Contributions.}
\begin{itemize} 
\item We introduce a new concept called buffer zones which is at the core of a new adversarial ML defense, coined BUZz. Using this BUZz defense, we are able to reduce the attack success rate for the strongest black-box adversary to as low as 7\%  on CIFAR-10 and to 11\% for FashionMNIST for the best known untargeted attack.
\item
We introduce a new metric called the $\delta$ value for comparing defenses.
While our defense does require some drop in clean accuracy (when compared to a vanilla scheme), we use $\delta$ to show that this is an acceptable trade-off for better security (this corresponds to a  smaller $\delta$).
\item
We show through rigorous experimentation with multiple untargeted attacks (black-box~\cite{Szegedy2014,PapernotMG16,AthalyeCarliniWagner2018,LiuChenLiuEtAl2016,PapernotMcDanielGoodfellowEtAl2017} based FGSM~\cite{GoodfellowShlensSzegedy2014}, BIM~\cite{kurakin2017adversarial}, MIM~\cite{dong2018boosting}, PGD~\cite{madry2018towards}, C$\&$W~\cite{carlini2017adversarial} and EAD~\cite{chen2018ead}) that
the BUZz defense is much better than what other well-known defenses in the literature can achieve. 
\item Our defense mechanism  provides a spectrum of possible choices between higher clean accuracy or better security. This allows a user of our defense to select the best trade-off for themselves, a choice previous unattainable by other defenses.
\end{itemize}

\textbf{Paper outline.} We organize the paper as follows: in Section~\ref{sec:background} we discuss adversarial black-box attacks.
%as well as how white-box attacks can be used in conjunction with them. 
In Section~\ref{sec:metric} we describe how the performance of a defense can be understood using a new intuitive metric that combines clean accuracy and the attack success rate. In Section~\ref{sec:methods} we explain how our defense based on buffer zones can be realized and the security arguments related to it. We present comprehensive experimental results for a wide array of attacks and defenses in Section~\ref{sec:experiment}. Finally we offer concluding remarks in Section~\ref{sec_conclusion}. 
\section{Preliminary knowledge}
\label{sec:background}
\noindent
\subsection{Adversarial examples in an image classification task}
The general scheme of a successful attack can be described as follows (see \cite{YuanHeZhuEtAl2017}). The adversary is given a trained image classifier (e.g, CNN classifier) $F$ which outputs a class label $l$ for a given input data (i.e., image) $x$. The adversary will add a perturbation $\eta$ to the original input $x$ to get an adversarial example (or a modified data input) $x'$, i.e., $x'=x+\eta$. Normally, $\eta$ should be small to make the adversarial noise barely recognizable by humans. Yet, the adversary may be able to fool the classifier $F$ to produce any desired class label $l' (\neq l)$. Assume that $f(x)=(s_1,s_2,\dotsc,s_k)$ is a $k$-dimensional vector of confidence scores $s_j$ of class labels $j$. We call $f(x)$ the score vector with $0\leq s_j\leq 1$, $\sum_{j=1}^k s_j =1$, and $k$ the number of class labels. % i.e., $f(x)=$ and
The class label $l$ is computed as
\begin{eqnarray*}
l &=& F(f(x)) %\\
 % &=& 
  = \underset{i\in [1,\dotsc,k]}{\text{argmax}} \{s_1,s_2,\dotsc,s_k\}.
\end{eqnarray*}
Given $x\in [0,1]^d$ and $l'\neq l=F(f(x))$, the attacker wishes to ideally solve %$\underset{x'\in [0,1]^d}{\min} \ 
$\min \{
\|x'-x \| \ : \ x'\in [0,1]^d\}$ such that $ F(f(x'))=l'\neq l=F(f(x))$,
where $l$ and $l'$ are the output labels of $x$ and $x'$,  $\|\cdot \|$ denotes the distance between two data samples, and $d$ is the number of dimensions of $x$. 
An untargeted attack means that the adversary is happy with any $l'\neq l$, while in a targeted attack the adversary specifies an adversarial label $l'\neq l$ a-priori.
%$\eta = x' - x$ is the perturbation added on $x$. 

\subsection{Black-box attacks} 
In this paper we operate under the assumption that the parameters of the classifier are secret, and focus on a black-box adversary. Black-box attacks use non-gradient information of classifier $F$ such as (part of) the original training data set $\mathcal{X}_0$~\cite{PapernotMG16} and/or a set $\mathcal{X}_1$ of adaptively chosen queries to $F$ (i.e., 
%$\{(x,F(f(x))) : x\in \mathcal{X}_1\}$ or 
$\{(x,l=F(f(x))) : x\in \mathcal{X}_1 \}$)~\cite{PapernotMcDanielGoodfellowEtAl2017} -- queries in $\mathcal{X}_1$ are not in the training data set $\mathcal{X}_0$. 
Based on information $\mathcal{X}_0$ and $\mathcal{X}_1$ the adversary trains its own copy of the proposed defense. This is an approximation of the defense and is called the synthetic model. Once the synthetic model has been created the adversary can run white-box attacks on the synthetic model to create adversarial examples. Then examples are submitted to the defense with the hope that the adversarial examples created to fool the synthetic model will also fool the defense model. In \cite{LiuChenLiuEtAl2016} it was shown that the transferability property of adversarial examples between different models which have the same topology/architecture and are trained over the same dataset is very high, i.e., nearly $100\%$ for  ImageNet \cite{RussakovskyDengSuEtAl2015}. This explains why the adversarial examples generated for the synthetic network can often be successful adversarial examples for the defense network. Black-box attacks can be partitioned into the following categories:
\noindent

\textbf{Pure black-box attack}~ \cite{Szegedy2014,PapernotMG16,AthalyeCarliniWagner2018,LiuChenLiuEtAl2016}: The adversary is \textit{only} given knowledge of a training data set $\mathcal{X}_0$. 
\noindent

\textbf{Oracle based black-box attack}~\cite{PapernotMcDanielGoodfellowEtAl2017}: The adversary is allowed to adaptively query target classifier $F$, which gives information $\mathcal{X}_1$.

    Based on this information, the adversary builds his own classifier $G$ which is used to produce adversarial examples using an existing white-box attack methodology. Compared to the native (pure) black box attack, this attack is supposed to be more efficient because $G$ is intentionally trained with examples labeled by $F$. Hence, the transferability between $F$ and $G$ may be significantly higher.   
\noindent

{\bf Mixed black-box attack}: In this paper we strengthen the original oracle based black-box attack~\cite{PapernotMcDanielGoodfellowEtAl2017} by doing the following: we allow the adversary access to the entire original training dataset and unlimited query access to the defense to train synthetic network $G$. Our mixed black-box attack is more powerful than both the pure black-box attack and oracle based black-box attack. We further confirm this fact through our experiments in Supplemental Material \ref{expana} (see Tables~\ref{tab:all_attack_all_defenses_CIFAR10} and~\ref{table:pureBB_Fashion_CIFAR}).

\noindent
\textbf{Zeroth order optimization based black-box attacks or score value vector based black-box attacks}~\cite{ChenPinYu2017}. The adversary does not build any assistant classifier $G$ as done in the previous black-box attacks. Instead, the adversary  adaptively queries $\{x,f(x),F(f(x))\}$ to approximate the gradient $\nabla f$ %(which cannot be directly computed due to lack of knowledge on the structure of $f$) 
    based on a derivative-free optimization approach. Using the approximated $\nabla f$, the adversary can build adversarial examples by directly working with %the classifier 
    $f$. Another attack in this line is called SimBA (Simple Black Box Attack)~\cite{guo2019simple}. This attack also requires the score vector $f(x)$ to mount the attack.   
    
\noindent
\textbf{Decision based black-box attack~\cite{Chen2019Hop}}. As shown in the literature, the adversary is still able to mount a black box attack if he is only allowed to access the decision value, i.e., $F(f(x))$. The main idea of the attack is trying to find the boundaries between the class regions using a binary searching methodology and gradient approximation for the points located on the boundaries. This type of attacks is called Boundary Attacks. As mentioned in~\cite{Chen2019Hop} Boundary Attacks are not as efficient as pure black box attacks since the adversary has to make many queries to create successful adversarial examples and this gives large query and computation complexities. 

%. Hence, if it would be very expensive in terms of query and computation complexities, if he wants to create many successful adversarial examples.

%\textcolor{red}{Ha: 
%As an important note, we want to 
We highlight that oracle based black-box attacks, mixed black-box attacks, score value vector based black-box attacks and decision based black-box attacks are called \textit{adaptive} black-box attacks as defined in~\cite{carlini2019}. In our adversarial model we consider an adversary with {\em only} black box access to labels $F(f(x))$ for chosen queries $x$. This assumes that the defender only offers classification to labels as a service (a reasonable assumption in practice) -- and this {\em excludes direct adversarial knowledge of score vectors $f(x)$}.  Given this adversarial model, we cover all different types of adaptive black-box attacks for our proposed defense BUZz: 
%We explain why the score value vector based black-box attacks do not work for our defense and. %
That is, we do not consider decision based black box attacks because its performance is not as good as the pure black box attack because of its increased query and computation complexities as mentioned above. 
%\textcolor{red}{As shown in~\cite{Chen2019Hop}, the adversary has to make many queries to create a successful adversarial examples. Hence, if it would be very expensive in terms of query and computation complexities, if he wants to create many successful adversarial examples.}
Therefore, we only need to consider pure black box attacks, oracle based black box attacks and mixed black box attacks. (More discussion is in Supplemental Material \ref{sec:apdx:A1}, \ref{app-white} and \ref{app-black}.)

For our defense based on the existence of buffer zones (see Figure~\ref{fig:fish}) any successful attack must generate a sufficient amount of noise to cross the buffer zone -- and since it is hard to find shallow parts in the buffer zone for crossing with visually imperceptible noise, we attain better security (a smaller attack success rate).\footnote{We hypothesise, but did not verify, that this problem remains difficult for an attacker even if s/he has been given access to the score vector $f(x)$ (implying that score value vector based attacks are also more difficult to pull off). We stress though that in our adversarial model the BUZz defense only outputs $F(f(x))$ and not $f(x)$.}

\section{Defense performance metric} \label{sec:metric}

We introduce a new metric to properly understand the combined effect of: 
\begin{enumerate}
    \item A drop $\gamma$ in clean accuracy from an original clean accuracy $p$  to clean accuracy
\begin{equation} p_d=p-\gamma \label{eq:gamma} \end{equation}
for the defense. Here, clean accuracy $p$ corresponds to a vanilla scheme without defense strategy in a non-malicious environment. 
Similarly, clean accuracy
$p_d$ represents the accuracy for the defense  measured in the non-malicious environment without adversaries. (We take  "clean" to have the additional meaning of being in a non-malicious environment.)
\item The attacker's success rate $\alpha$ against the defense. If the defense recognizes an adversarial manipulated image as an adversarial example, then it outputs the adversarial label $\bot$ and the attack is not considered successful. When defining $\alpha$, we restrict ourselves to adversarial examples for those images which the defense  
(in their original non-attacked form) properly classifies by their correct labels. The attacker's success rate is then defined as the fraction of adversarial examples that manipulate these images 
in such a way that the defense produces labels different from the correct labels and different from the adversarial label $\bot$.
For completeness, literature defines the defense accuracy or defense success rate as $1-\alpha$ (most defenses cannot recognize an adversarial manipulated image as an adversarial example and do not have an adversarial label as possible output).
\end{enumerate}
%So, in a non-malicious environment we have the original clean accuracy $p$ while in the malicious environment 

Proper/Accurate classification by the defense in the presence of adversaries is one of the following: An image (possibly after adversarial manipulation) is recognized by its correct label (implying no attack was present or the attack did not work). Or, an adversarial manipulated image is given the (correct) adversarial label $\bot$ (if the defense offers this possibility). 

The probability of proper/accurate classification by the defense in the presence of adversaries is equal to $(p-\gamma)(1-\alpha)$ (since the defense properly labels a fraction $p-\gamma$ if no adversary is present and out of these images a fraction $\alpha$ is successfully attacked if an adversary is present).
In other words $(p-\gamma)(1-\alpha)$ is the
accuracy of the defense in the presence of adversaries (malicious environment).
Going from a non-malicious environment without defense to a malicious environment with defense gives a drop in  accuracy of
\begin{align}
\label{eq_important}
\delta = p - (p-\gamma)(1-\alpha) = \gamma + (p-\gamma)\alpha.
\end{align}
%In order to minimize this drop $\delta$, it turns out to be very important to have $\alpha$ small enough, which is accomplished in this paper. 
$\delta$ can be used to measure the effectiveness of different defenses, the smaller the better. If two defenses offer roughly the same $\delta$, then it makes sense to consider their $(\gamma,\alpha)$ pairs and choose the defense that either has the smaller $\alpha$ or the smaller $\gamma$.

We may use subscript $t$ and $u$ in $\delta_t$ and $\delta_u$ when differentiating for targeted attacks
%\footnote{The definition of $\delta_t$ assumes into some extent a worst-case scenario for the defender: Even though $(p-\gamma)(1-\alpha)$ represents the accuracy of the defense under best known targeted attacks, this does not mean that a fraction $1-(p-\gamma)(1-\alpha)$ will } 
and untargeted attacks. Since untargeted attacks are easier to pull off, $\delta_t\leq \delta_u$. In this paper we focus on defending against the easier untargeted attacks and use $\delta=\delta_u$. I.e., we also want to defend against the range of attacks that are easier for an attacker  to have success with. 

From a pure ML perspective, in order for a defense to perform well in a non-malicious environment, we want $\gamma$ very small or, equivalently, $p_d$ close to $p$. From a pure security perspective, in order for a defense to perform well in a malicious environment, we want $\delta$ to be small. Therefore, for properly comparing defenses we focus on tuples $(\delta= \gamma + (p-\gamma)\alpha, p_d=p-\gamma)$, 
%in Table~\ref{tab:all_attack_all_defenses_CIFAR10}, 
where $\alpha$ corresponds to the best attacker's success rate across the best known (untargeted, respectively targeted) attacks from literature. Notice that the vanilla scheme can be considered in a malicious environment as well and this will correspond to some $( \delta_{van}, p_d=p)$. Clearly defenses that result in $\delta\geq \delta_{van}$ do not improve over implementing no defense at all (which is the plain vanilla scheme). 

It turns out that having both $\gamma$ and $\delta$ small is difficult to achieve: See (\ref{eq_important}), the smaller $\gamma$, the more $\delta$ behaves like $p\alpha$ and as of now we do not know how to achieve a very small attacker's success rate $\alpha$, say less than a couple percent, while keeping $\gamma$ very small. In this paper we show $\alpha <11\%$ 
%($=1-p_d$ in Table~\ref{tab:all_attack_all_defenses_CIFAR10}) 
for defenses based on buffer zones -- this is significantly lower than what other state-of-the-art defenses offer, but is at the cost of a higher $\gamma$.
%(more than just a couple $2-3\%$).

\section{Methodology}
\label{sec:methods}
The BUZz defense is based on the concept of buffer zones. Buffer zones are the regions in between classes where if an input falls in this region, it is marked as adversarial. In theory, this forces the adversary to add noise $\eta$ greater than a certain magnitude in order to overcome (cross over) the buffer zone. Because an attack fails if the noise becomes visual perceptible to humans, the adversary is limited in terms of the magnitude of $\eta$. In many cases this means the adversary may not be able to overcome the buffer zone and therefore cannot fool the classifier. The natural question is how can buffer zones be implemented in classifiers? In this section we discuss different techniques that can be used to create buffer zones.
\subsection{Realizing buffer zones through thresholding}
For a given single classifier or network, we can create buffer zones between the regions of class labels based on a threshold mechanism. Assume $f(x)=(s_1,...,s_k)$ as the vector of confidence scores where $s_i$ is the score of class labels $j$ for a given classifier $f$ and input $x$. The corresponding predicted class label $l$ is different from $\bot$ if there exists at least one $s_i$ such as $s_i \geq \theta$ where $\theta \in [0,1]$. 

\noindent
\textbf{Security argument.} Intuitively, the buffer zones between the regions are created because of the following reason. Assume that the input $x$ locates at the center of a region with class label $0$ as in Figure~\ref{fig:fish}. %fig:key_idea}. 
Then the score value of $s_0$ should be very high, i.e., close to 1. Score $s_0$ drops when $x$ moves towards the boundary between regions having class labels $0$ and $1$, while $s_{1}$ increases. Score $s_{1}$ will be highest when $x$ reaches the center of the region with class label $1$. Therefore, using threshold $\theta$, we effectively create buffer zones between the regions. It implies that the adversary must create a sufficient large noise $\eta$ to make $x$ jump from the region with class label $0$ to the region with class label $1$ across the buffer zones between them.   

\textbf{Advantages and disadvantages.} The advantage of the threshold approach is that it is very simple, efficient and straightforward to implement. It does not require any image transforms or classifier retraining. However, there are drawbacks to this approach. For certain datasets with low complexity (see Supplemental Material \ref{expana} for Fashion-MNIST in Table~\ref{table:mixBB_Fashion_all_BUZz}) this technique offers only limited security, even when the threshold is set as high as $0.99$. Furthermore, if only a single network is involved, then other black-box attacks may  be developed where the focus is merely on producing high confidence adversarial examples using the synthetic network.  
\subsection{Realizing buffer zones through multiple networks}
\begin{figure}[htp!]
\begin{center}
\includegraphics[width = 0.3\textwidth]{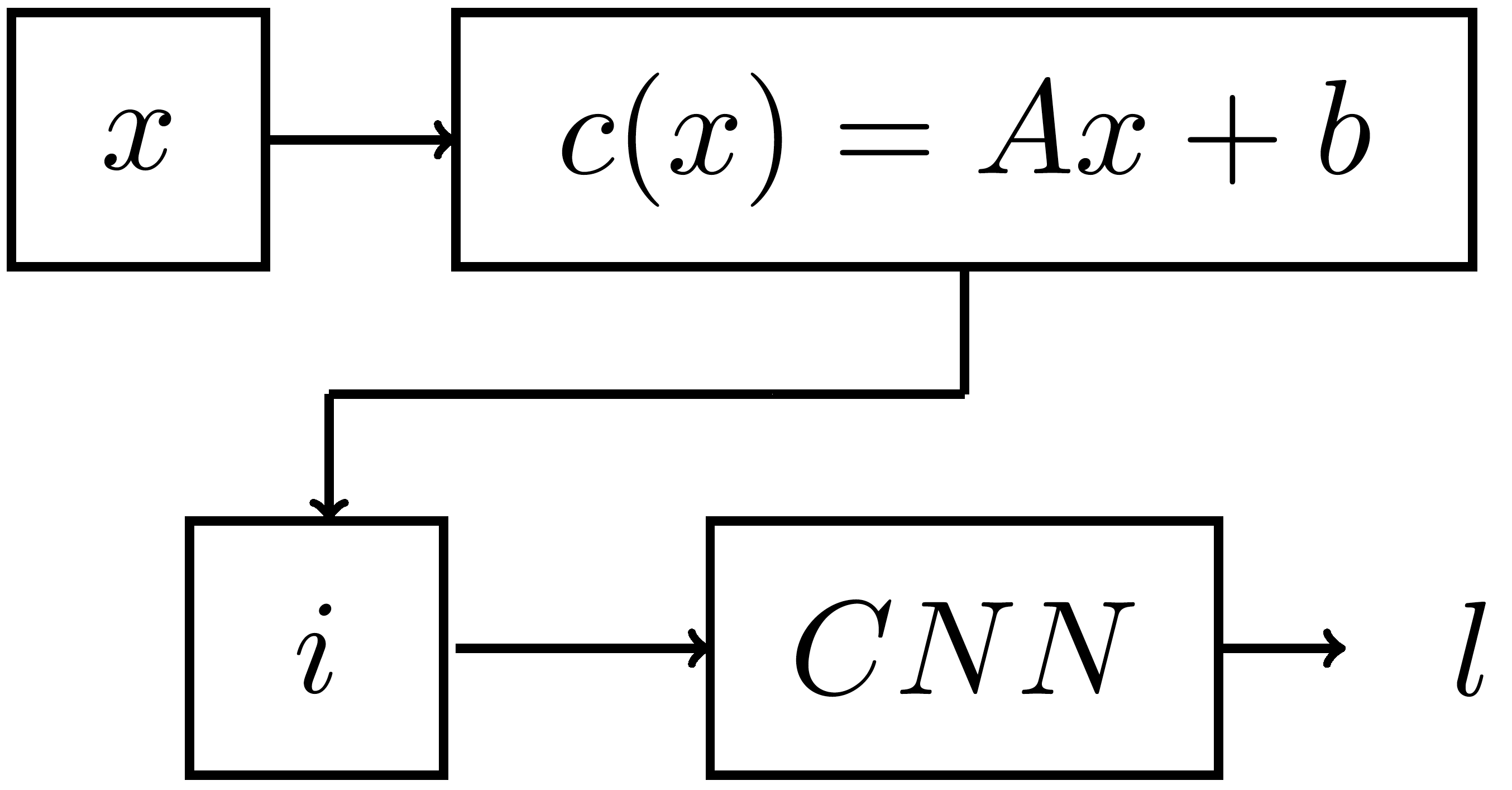}
\includegraphics[width = 0.6\textwidth]{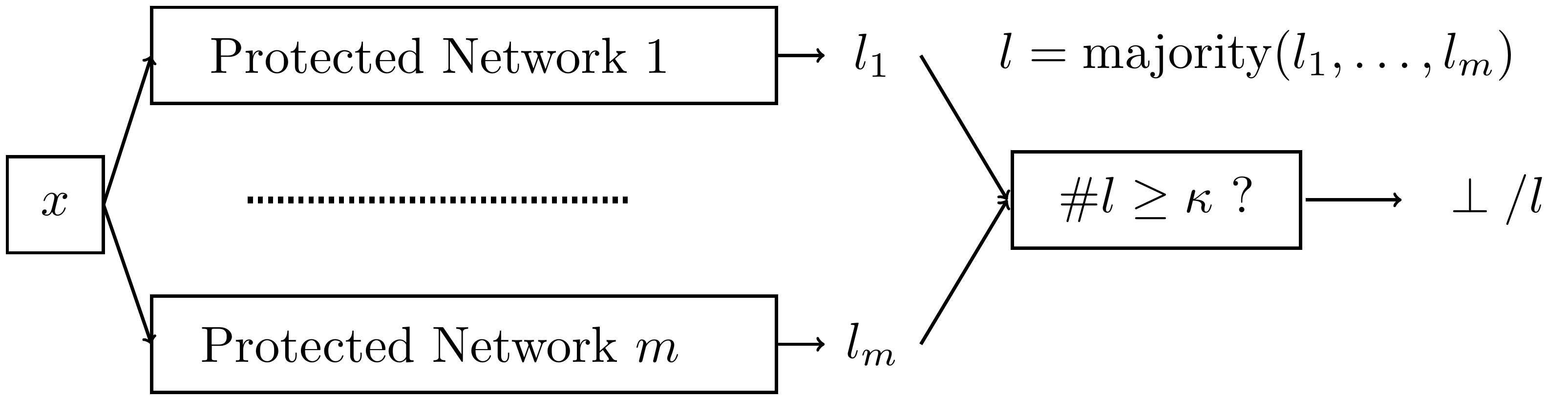}
\caption{Design of protected network (left) and BUZz: majority voting and threshold $\kappa$ (right).}\label{fig:buzz}
\end{center}
\end{figure}

Buffer zones can be created through the use of multiple networks. A naïve approach to this method would be to simply use networks with different architectures. However, we experimentally show that merely using different architectures does not yield security. This has also been shown in the literature in~\cite{LiuChenLiuEtAl2016}. To break transferability between networks we introduce secret image transformations for each classifier. Our defense is composed of multiple classifiers is depicted in Figure~\ref{fig:buzz} (right). Each CNN has two {\em unique image transformations} as shown in Figure~\ref{fig:buzz} (left). The first is a fixed randomized linear transformation $c(x)=Ax+b$, where $A$ is a matrix and $b$ is a vector. 

After the linear transformation a resizing operation $i$ is applied to the image before it is fed into the CNN. The CNN corresponding to $c$ and $i$ is trained on clean data $\{i(c(x))\}$. This results in a weight vector $w$. The $m$ protected layers in Figure~\ref{fig:buzz} are described by `parameters' $(c_j, i_j, w_j)_{j=1}^m$.

% \begin{figure}[!h]
% \centering
% \includegraphics[scale=0.33]{./Figs/tikz/secret_random_layer16.pdf}
% \caption[]{Design of protected network.}
%  \label{fig:enhance_design_a}
% %\vspace{2mm}
% \noindent
% \end{figure}

% \begin{figure}[!h]
% \centering
% \includegraphics[scale=0.23]{./Figs/tikz/secret_random_layer17.pdf}
% \caption[]{BUZz: majority voting and threshold $\kappa$.}
%  \label{fig:general_defense_b}
% %\vspace{2mm}
% \noindent
% \end{figure}

When a user wants to query the defense, input $x$ is submitted to each protected network which computes its corresponding image transformation and executes its CNN. The outputs of the protected networks are class labels $(l_j)_{j=1}^m$. The final class label of BUZz, i.e., the composition of the $m$ protected networks, is a majority vote based on a threshold $\kappa$. 
In our experiments we use %The output of each CNN is then computed and the final class label is based on 
unanimous voting, i.e., if the networks do not all output the same class label then the adversarial/undetermined class label $\bot$ is given as the output (i.e., $\kappa=m$).
\noindent

\textbf{Security argument.} 
In a multi-classifier defense, buffer zones can be established by using majority voting and diminishing the transferability among the different classifiers through image transformations. To decrease the transferability, we can train each classifier on a different transformed input $x$. From~\cite{GuoRanaCisseEtAl2017} we know adversarial examples are sensitive to image transformations which either distort the value of the pixels in the image or change the original spatial location of the pixels. 

In the context of Papernot's oracle based black-box attack and pure black-box attack, the adversarial noise $\eta$ is created based on a white-box attack for a synthetic network (score vector) $g$ of BUZz. It means that the noise $\eta$ is specifically developed for $g$. Since the $x'=x+\eta$ is input into every protected network of BUZz, the $j$-th layer will apply its CNN network on a noisy image $i_j(c_j(x'))=i_j(c_j(x+\eta))$, which due to the linearity of $i_j(c_j(\cdot))$ is equal to $i_j(c_j(x))+i_j(c_j(\eta))$. Therefore, networks receive different input noise $i_j(c_j(\eta)) \neq \eta$. Since the transformations are different for each of the classifiers that participate in the majority vote, the attack success rate can be reduced. It is important to note that in this paper we have experimentally established that image resizing and linear transformations can reduce the transferability. However, there may be other image transformations that can also accomplish this goal. Notice though that it has cost us significant time and effort to find the proposed transformations since the drop 
$\gamma$ in clean accuracy should be designed as small as possible. 

\noindent
\textbf{Advantages and disadvantages.} The advantage of BUZz using multiple networks and image transformations is that it can achieve a higher defense accuracy for certain datasets (we show this experimentally in Supplemental Material \ref{expana} for Fashion-MNIST in Table \ref{table:mixBB_Fashion_all_BUZz}) than only using thresholding. We also have more control over the tradeoff between the attacker's success rate and clean accuracy using the linear transformation $Ax+b$. Here $A$ is a fixed $n$ by $n$ matrix and $b$ a fixed $n$ by $n$ bias. For example, if only $b$ is random (with small magnitude) and $A$ is identity, it results in less image distortion (so higher clean accuracy) but also less security (more adversarial samples bypass the defense). The disadvantage of this approach is that each network must be retrained to handle a certain image transformation. 

\subsection{Realizing buffer zones through combinational approaches}
Previously we discussed how buffer zones can be realized through thresholding or through multiple networks with image transformations. A natural extension of these two techniques is a combinational approach. In this type of defense multiple networks with image transformations are employed, some of which also implement thresholding. In this way we can reduce the number of classifiers that need to be retrained. At the same time in this defense we still have the extra flexibility that the linear transformation provides in choosing between security and clean accuracy. We experiment with this combinational approach, %in 
%Section~\ref{sec:experiment} 
e.g., BUZz2 has two classifiers of which one uses a threshold 0.70 (BT2-70).

It is important to note that it may be possible to further combine other defense techniques such as adversarial retraining~\cite{TramerKurakinPapernotEtAl2017} with BUZz. However, the goal of this paper is not to exhaustively test every defense combination but merely propose and experimentally verify a working defense framework.

\section{Experiments}
\label{sec:experiment}

In this section we provide experimental results to show the effectiveness of the BUZz defense. We experiment with two popular datasets, Fashion-MNIST~\cite{Han2017FashionMNIST} and CIFAR-10~\cite{CIFAR10ref}. Fashion-MNIST is a 10-class dataset that has 60,000 training and 10,000 testing images in grayscale. CIFAR-10 is also a 10-class dataset with 50,000 training and 10,000 testing images in color. All images are normalized in the range [0,1] with a shift of -0.5 so that they are in the range [-0.5, 0.5]. In terms of network architecture, we use ResNet56~\cite{he2016deep} for the networks in the CIFAR-10 defenses and VGG16~\cite{simonyan2014very} for the networks in the Fashion-MNIST defenses. Full architectural details as well as their standard training procedure can be found in  Supplemental Material \ref{app:pseudo_alg}. 

\subsection{Defenses}
{\bf BUZz Defenses.} For the BUZz defense, we experimented with three different variants. We tested buffer zones based on image transformations, buffer zones based on thresholding and buffer zones based on a combination of image transformations and thresholding. For BUZz based on image transformations each network has an image transformation selected from mappings $c(x)= Ax + b$. We explain how we chose the randomized $A$ and $b$ based on the dataset in  Supplementary Material \ref{app:pseudo_alg}. We can think of an image transformation $c_j(x)$ as an extra randomly fixed layer added to the layers which form the $j$-th CNN. We tested three of these designs: One with 8 networks each using a different image resizing operation from 32 to 32, 40, 48, 64, 72, 80, 96, 104. The second with 4 networks being the subset of the 8 networks that use image resizing operations from 32 to 32, 48, 72, 96. The third with 2 networks being a subset of the 8 networks that use image resizing operations from 32 to 32 and 104.

For buffer zones based on thresholding we tested the following BUZz thresholding configurations: a single vanilla network with thresholding cutoff 0.7, thresholding cutoff 0.95 and thresholding cutoff 0.95. For the combination of the two approaches we tested BUZz-2 with thresholding: each network has the mapping $c(x)= Ax + b$, one of the networks has a resize operation 32 (no resize), the second network has a resize operation 104 and a thresholding cutoff (0.7 for CIFAR-10 and 0.95 for Fashion-MNIST).  
\noindent

{\bf Literature defenses.} In addition to our buffer zone based defense, we also test other defenses from the literature. While it is impossible to exhaustively test all published defenses, here we focus on a selection of the most prominent ones including
\cite{XieWangZhangEtAl2017,LiuChenLiuEtAl2016,CaoGong2017,cohen2019certified,GuoRanaCisseEtAl2017,TramerKurakinPapernotEtAl2017,srisakaokul2018muldef}.
%Xie's: Randomized 1-Net~\cite{XieWangZhangEtAl2017}, Liu's: Mixed Arch 2-Net~\cite{LiuChenLiuEtAl2016}, Cao's: randomized smooth technique~\cite{CaoGong2017,cohen2019certified},  Guo's: this is BUZz with a single network~\cite{GuoRanaCisseEtAl2017}, Tramer's: adversarial training~\cite{TramerKurakinPapernotEtAl2017}, MD2 and MD4~\cite{srisakaokul2018muldef}.  
See Supplemental Material \ref{literature} for a discussion on a more extensive list of defenses together with experimental results.

\subsection {Attacks}
%The hierarchy of 
The black-box attacks we experiment with can be divided into two different types. The first type is generation of the synthetic model using training data and synthetic data labeled by the oracle~\cite{PapernotMcDanielGoodfellowEtAl2017} which we call the mixed black-box attack. 
%The parameters for this attack are given in Table~\ref{table:setup_datasets}. 
The second type of black-box attack we consider is the pure black-box attack. In this attack the generation of the synthetic model is accomplished using the original training dataset and original training labels. 
%The parameters for this attack are given in Table~\ref{table:setup_datasets}. 
In both black-box attacks after the synthetic model is generated, we can run any white-box attack on the synthetic model to create adversarial samples to try and fool the defense. The white-box attacks we consider are FGSM~\cite{goodfellow2014explaining}, BIM~\cite{kurakin2017adversarial}, MIM~\cite{dong2018boosting}, PGD~\cite{madry2018towards}, Carlini$\&$Wagner\cite{carlini2017adversarial} and EAD~\cite{chen2018ead} (see Supplemental Material \ref{app-white}). %The parameters for these attacks are given in Table~\ref{tab:attack_parameters}.    

%\begin{table}[th!]
%\caption{Attacks' parameters. $i$ - number of iterations, $d$ - decaying factor, $r$ radius of the ball for generating the initial noise, $c$ - constant value of C$\&$W attack, $\epsilon$ - noise magnitude, $\beta$ - constant value of EAD attack. Binary Search = Bi.Sr}
%\centering
%\scalebox{0.72}{
%\begin{tabular}{|p{1cm}|p{4.5cm}|p{4.5cm}| }
%\hline  
%Attacks & Fashion-MNIST              & CIFAR-10 \\ \hline \hline
%FGSM    & $\epsilon = 0.15$          & $\epsilon=0.05$  \\ \hline 
%BIM     & $i=10,\epsilon=0.015$      & $i=10,\epsilon=0.005$ \\ \hline
%PGD     & $i=10,r=0.031,\epsilon=0.015$      & $i=10,r=0.031,\epsilon=0.005$ \\ \hline
%MIM     & $i=10,d=1.0,\epsilon=0.015$      & $i=10,d=1.0,\epsilon=0.005$ \\ \hline
%C$\&$W  & $i=1000,c=\text{Bi.Sr}$      & $i=1000,c=\text{Bi.Sr}$ \\ \hline
%EAD  & $i=1000,c=\text{Bi.Sr},\beta=0.01$      & $i=1000,c=\text{Bi.Sr},\beta=0.01$ \\ \hline
%\hline
%\end{tabular}
%}
%\label{tab:attack_parameters}
%\end{table}
%%table* for 2 columns

% ==================== (MOVE - END) =======================

\subsection{Experimental analysis}

%In Table~\ref{table:pureBB_Fashion_CIFAR} we show the results for the pure black-box attack on CIFAR-10 and Fashion-MNIST for various BUZz defenses. Here we do not test every defense as these experimental results are only used to demonstrate two points. 

We first observe that experiments show that the pure black-box attack is always equal to or weaker than the mixed black-box attack.
%%if you compare these table results to those in    Table~\ref{table:mixBB_Fashion_all_BUZz} and Table~\ref{tab:all_attack_all_defenses_CIFAR10}. 
%Furthermore, the BUZz defense for CIFAR-10 and Fashion-MNIST is able to mitigate this type of attack, in particular, the defense accuracy of BUZz-8 is $88\%$ or greater for all attacks.
%
%In Table~\ref{table:mixBB_Fashion_all_BUZz} we show the results for the mixed black-box attack on Fashion-MNIST for various BUZz defense configurations and a vanilla (no defense) network. In this table we also show the clean accuracy and $\delta$ metric (the best $\delta$ is obtained by the BUZz-8 defense). We can interpret Table~\ref{table:mixBB_Fashion_all_BUZz} by figure \ref{fig:key_idea3}.
%
%% FASHION MNIST TO APPENDIX
%\begin{figure}%[!h]
%\centering
%\includegraphics[scale=0.25]{./Figs/tikz/defensegraph3.pdf}
%\caption[]{Fashion-MNIST defenses with clean accuracy $p_d$ (Clean Accuracy) and $\delta$ value (Delta) for untargeted attacks.}
% \label{fig:key_idea3}
%%\vspace{2mm}
%\noindent
%\end{figure}
%\noindent 
%
%In Table~\ref{tab:all_attack_all_defenses_CIFAR10}, 
For CIFAR-10, Figure \ref{fig:key_idea} plots for various defenses the clean accuracy $p_d$ and $\delta$ value for best  untargeted black-box attacks among the ones listed above (MIM is generally the best performing). The defenses considered are BUZz 2, 4, 8, BUZz using threshold cutoff 0.7, 0.95 and 0.99, the combination BT2-70, and plain classifier (Vanilla), Xie's: Randomized 1-Net~\cite{XieWangZhangEtAl2017}, Liu's: Mixed Arch 2-Net~\cite{LiuChenLiuEtAl2016}, Cao's: randomized smooth technique~\cite{CaoGong2017,cohen2019certified},  Guo's: this is BUZz with a single network~\cite{GuoRanaCisseEtAl2017}, Tramer's: adversarial training~\cite{TramerKurakinPapernotEtAl2017}, MD2 and MD4~\cite{srisakaokul2018muldef}. 
For example BUZz2 achieves  $p_d=0.85$ and defender success rate $1-\alpha= 0.68$, which yields, by using $p=0.93$ for the vanilla scheme, $\delta = 0.93 - 0.85\times 0.68 = 0.36$ and $\gamma = 0.93 - 0.85 = 0.07$. 
We see that for the broader class of untargeted attacks BUZz offers a significantly better defense when compared to our selection of  most prominent defenses in literature. 
Minimal $\delta=0.22$ or $0.23$ is achieved for BUZz4 with $\alpha=0.14$ and $\gamma=0.12$ (top left corner); BT99 with $\alpha=0.11$ and $\gamma=0.14$; BUZz8 with $\alpha=0.07$ and $\gamma=0.17$. Among these BUZz4 maximizes clean accuracy $p_d=p-\gamma$. Notice that the smallest $\alpha$ in literature by Liu~\cite{LiuChenLiuEtAl2016} is $0.37$, significantly higher than for BUZz (although Liu only has a drop $\gamma=0.07$; in this case our BT95 achieves lower $\alpha=0.23$ for the same $\gamma=0.07$).
Supplemental Material \ref{expana} provides extensive details in the form of tables for our experiments on CIFAR-10 as well as Fashion-MNIST (for both untargeted and targeted attacks). For Fashion-MNIST we have $p=0.94$ with BUZz4 giving $\delta=0.26$ with $\alpha=0.11$ and $\gamma=0.17$.

\section{Conclusion}\label{sec_conclusion}

We introduced a new concept called buffer zones which is at the core of a new adversarial ML defense, coined BUZz. Using this BUZz defense, we are able to reduce the attack success rate to as low as $7\%$ for the strongest black-box adversary on CIFAR-10 and to $11\%$ for Fashion-MNIST. This methodology offers defense accuracy much better than other well-known defenses in the literature which we show through rigorous experimentation with multiple attacks (black-box based FGSM, BIM, MIM, PGD, C$\&$W and EAD, where we introduced the stronger mixed black-box attack). While our defense does require some drop in clean accuracy, we have shown that this is an acceptable trade-off for better security using a new metric called the $\delta$ value. The metric is used to compare the clean accuracy of a vanilla scheme without adversaries, i.e., {\em non-malicious environment without defense},  versus the  accuracy of a scheme with defense in the presence of adversaries, i.e., {\em malicious environment with defense}. In addition, we are the first to provide a spectrum of possible choices between higher clean accuracy or higher security. This allows a user of our defense to select the best trade-off for themselves, a choice previous unattainable by other defenses.

\newpage

\section*{Broader Impact Statement}

Machine learning has shown to be of crucial importance in data analytics. During the last decade deep learning and reinforcement learning together with multicore and GPU based computing have been able to push the field forward. Classification models can be learned with better and better clean accuracies. A central question, however, is whether the learned models can be fooled by adversaries. 
Adversarial machine learning -- and, in particular, black-box attacks on classification models -- is a serious threat to economy and safety. If adversarial examples are successful, then this may jeopardize  the safety offered by applications that use
%our US homeland security through surveillance, other kinds of 
image recognition software, etc. 
Research in adversarial machine learning aims at creating a solid understanding and theory  on how adversarial examples can be generated and how defense mechanisms can be designed to filter out adversarial examples or to still label them correctly. The proposed $\delta$-metric allows proper comparison of defense mechanisms across clean accuracy and attacker's success rate trade-offs -- the resulting benchmarking will help the scientific community to focus on promising defense heuristics and discard those which clearly do not help reducing $\delta$. Within this perspective, the paper also introduces buffer zones as a new technique for creating defenses that significantly reduce $\delta$. 
%We expect the scientific community to pick up on this research direction.
%The proposal introduces the key idea of buffer zones which will be further explored if this proposal gets awarded; by helping our scientific community through publishing our attack and defense software we expect to create a new niche of research on buffer zones with many other research groups participating.

%Our proposal offers a fundamental step forward to offering much better security than what  state-of-the-literature can give. One of the proposed tasks is to build a proof-of-concept registration service with image classification software which exploits a lower $\delta$ so that registration of an object using multiple images can only with small probability be impersonated with adversarial examples that do not fool the human eye. This will help in understanding how third party classification can become trusted in providing better security and safety. 

% \clearpage
% \small
\bibliography{all_refs,ref2}

\begin{thebibliography}{45}
\providecommand{\natexlab}[1]{#1}
\providecommand{\url}[1]{\texttt{#1}}
\expandafter\ifx\csname urlstyle\endcsname\relax
  \providecommand{\doi}[1]{doi: #1}\else
  \providecommand{\doi}{doi: \begingroup \urlstyle{rm}\Url}\fi

\bibitem[Athalye et~al.(2018{\natexlab{a}})Athalye, Carlini, and
  Wagner]{AthalyeCarliniWagner2018}
Anish Athalye, Nicholas Carlini, and David~A. Wagner.
\newblock {Obfuscated Gradients Give a False Sense of Security: Circumventing
  Defenses to Adversarial Examples}.
\newblock In \emph{ICML 2018}, pages 274--283, 2018{\natexlab{a}}.

\bibitem[Athalye et~al.(2018{\natexlab{b}})Athalye, Engstrom, Ilyas, and
  Kwok]{AthalyeEngstromIlyasEtAl2018}
Anish Athalye, Logan Engstrom, Andrew Ilyas, and Kevin Kwok.
\newblock {Synthesizing Robust Adversarial Examples}.
\newblock In \emph{ICML 2018}, pages 284--293, 2018{\natexlab{b}}.

\bibitem[Cao and Gong(2017)]{CaoGong2017}
Xiaoyu Cao and Neil~Zhenqiang Gong.
\newblock {Mitigating Evasion Attacks to Deep Neural Networks via Region-based
  Classification}.
\newblock \emph{CoRR}, abs/1709.05583, 2017.
\newblock URL \url{http://arxiv.org/abs/1709.05583}.

\bibitem[Carlini and Wagner(2017{\natexlab{a}})]{carlini2017adversarial}
Nicholas Carlini and David Wagner.
\newblock {Adversarial examples are not easily detected: Bypassing ten
  detection methods}.
\newblock In \emph{AISec@CCS}, 2017{\natexlab{a}}.

\bibitem[Carlini and Wagner(2016)]{CarliniWagner2016}
Nicholas Carlini and David~A. Wagner.
\newblock {Towards Evaluating the Robustness of Neural Networks}.
\newblock \emph{CoRR}, 2016.

\bibitem[Carlini and Wagner(2017{\natexlab{b}})]{CarliniWagner2017}
Nicholas Carlini and David~A. Wagner.
\newblock {MagNet and "Efficient Defenses Against Adversarial Attacks" are Not
  Robust to Adversarial Examples}.
\newblock \emph{CoRR}, abs/1711.08478, 2017{\natexlab{b}}.

\bibitem[Carlini et~al.(2019)Carlini, Athalye, Papernot, Brendel, Rauber,
  Tsipras, Goodfellow, Madry, and Kurakin]{carlini2019}
Nicholas Carlini, Anish Athalye, Nicolas Papernot, Wieland Brendel, Jonas
  Rauber, Dimitris Tsipras, Ian~J. Goodfellow, Aleksander Madry, and Alexey
  Kurakin.
\newblock On evaluating adversarial robustness.
\newblock \emph{CoRR}, abs/1902.06705, 2019.

\bibitem[Chen and Jordan(2019)]{Chen2019Hop}
Jianbo Chen and Michael~I. Jordan.
\newblock Boundary attack++: Query-efficient decision-based adversarial attack.
\newblock \emph{CoRR}, abs/1904.02144, 2019.
\newblock URL \url{http://arxiv.org/abs/1904.02144}.

\bibitem[Chen et~al.(2017)Chen, Zhang, Sharma, Yi, and Hsieh]{ChenPinYu2017}
Pin-Yu Chen, Huan Zhang, Yash Sharma, Jinfeng Yi, and Cho-Jui Hsieh.
\newblock {ZOO: Zeroth Order Optimization Based Black-box Attacks to Deep
  Neural Networks Without Training Substitute Models}.
\newblock In \emph{AISec}. ACM, 2017.

\bibitem[Chen et~al.(2018)Chen, Sharma, Zhang, Yi, and Hsieh]{chen2018ead}
Pin-Yu Chen, Yash Sharma, Huan Zhang, Jinfeng Yi, and Cho-Jui Hsieh.
\newblock Ead: elastic-net attacks to deep neural networks via adversarial
  examples.
\newblock In \emph{Thirty-second AAAI conference on artificial intelligence},
  2018.

\bibitem[Cohen et~al.(2019)Cohen, Rosenfeld, and Kolter]{cohen2019certified}
Jeremy Cohen, Elan Rosenfeld, and Zico Kolter.
\newblock Certified adversarial robustness via randomized smoothing.
\newblock In \emph{International Conference on Machine Learning}, pages
  1310--1320, 2019.

\bibitem[Dong et~al.(2018)Dong, Liao, Pang, Su, Zhu, Hu, and
  Li]{dong2018boosting}
Yinpeng Dong, Fangzhou Liao, Tianyu Pang, Hang Su, Jun Zhu, Xiaolin Hu, and
  Jianguo Li.
\newblock Boosting adversarial attacks with momentum.
\newblock In \emph{Proceedings of the IEEE conference on computer vision and
  pattern recognition (CVPR)}, pages 9185--9193, 2018.

\bibitem[Feinman et~al.(2017)Feinman, Curtin, Shintre, and
  Gardner]{FeinmanCurtinShintreEtAl2017}
Reuben Feinman, Ryan~R. Curtin, Saurabh Shintre, and Andrew~B. Gardner.
\newblock {Detecting Adversarial Samples from Artifacts}.
\newblock \emph{CoRR}, abs/1703.00410, 2017.

\bibitem[Girshick(2015)]{Girshick2015}
Ross~B. Girshick.
\newblock {Fast {R-CNN}}.
\newblock In \emph{IEEE-ICCV}, pages 1440--1448, 2015.

\bibitem[Goodfellow et~al.(2014)Goodfellow, Shlens, and
  Szegedy]{GoodfellowShlensSzegedy2014}
Ian~J. Goodfellow, Jonathon Shlens, and Christian Szegedy.
\newblock {Explaining and Harnessing Adversarial Examples}.
\newblock \emph{CoRR}, abs/1412.6572, 2014.
\newblock URL \url{http://arxiv.org/abs/1412.6572}.

\bibitem[Goodfellow et~al.(2015)Goodfellow, Shlens, and
  Szegedy]{goodfellow2014explaining}
Ian~J Goodfellow, Jonathon Shlens, and Christian Szegedy.
\newblock Explaining and harnessing adversarial examples.
\newblock \emph{International Con- ference on Learning Representations (ICLR)},
  2015.

\bibitem[Guo et~al.(2017)Guo, Rana, Ciss{\'{e}}, and van~der
  Maaten]{GuoRanaCisseEtAl2017}
Chuan Guo, Mayank Rana, Moustapha Ciss{\'{e}}, and Laurens van~der Maaten.
\newblock {Countering Adversarial Images using Input Transformations}.
\newblock \emph{CoRR}, 2017.

\bibitem[Guo et~al.(2019)Guo, Gardner, You, Wilson, and
  Weinberger]{guo2019simple}
Chuan Guo, Jacob~R Gardner, Yurong You, Andrew~Gordon Wilson, and Kilian~Q
  Weinberger.
\newblock Simple black-box adversarial attacks.
\newblock \emph{arXiv preprint arXiv:1905.07121}, 2019.

\bibitem[He et~al.(2016)He, Zhang, Ren, and Sun]{he2016deep}
Kaiming He, Xiangyu Zhang, Shaoqing Ren, and Jian Sun.
\newblock Deep residual learning for image recognition.
\newblock In \emph{Proceedings of the IEEE conference on computer vision and
  pattern recognition}, pages 770--778, 2016.

\bibitem[Kingma and Ba(2014)]{kingma2014adam}
Diederik~P Kingma and Jimmy Ba.
\newblock Adam: A method for stochastic optimization.
\newblock \emph{arXiv preprint arXiv:1412.6980}, 2014.

\bibitem[Krizhevsky et~al.()Krizhevsky, Nair, and Hinton]{CIFAR10ref}
Alex Krizhevsky, Vinod Nair, and Geoffrey Hinton.
\newblock Cifar-10 (canadian institute for advanced research).
\newblock URL \url{http://www.cs.toronto.edu/~kriz/cifar.html}.

\bibitem[Krizhevsky et~al.(2012)Krizhevsky, Sutskever, and
  Hinton]{KrizhevskySutskeverHinton2012}
Alex Krizhevsky, Ilya Sutskever, and Geoffrey~E Hinton.
\newblock {ImageNet Classification with Deep Convolutional Neural Networks}.
\newblock In F.~Pereira, C.~J.~C. Burges, L.~Bottou, and K.~Q. Weinberger,
  editors, \emph{NIPS}, pages 1097--1105. 2012.

\bibitem[Kurakin et~al.(2016)Kurakin, Goodfellow, and
  Bengio]{KurakinGoodfellowBengio2016}
Alexey Kurakin, Ian~J. Goodfellow, and Samy Bengio.
\newblock {Adversarial Machine Learning at Scale}.
\newblock \emph{CoRR}, abs/1611.01236, 2016.
\newblock URL \url{http://arxiv.org/abs/1611.01236}.

\bibitem[Kurakin et~al.(2017)Kurakin, Goodfellow, and
  Bengio]{kurakin2017adversarial}
Alexey Kurakin, Ian Goodfellow, and Samy Bengio.
\newblock Adversarial examples in the physical world.
\newblock \emph{International Conference on Learning Representations (ICLR)
  Workshop}, 2017.

\bibitem[Liu et~al.(2017)Liu, Chen, Liu, and Song]{LiuChenLiuEtAl2016}
Yanpei Liu, Xinyun Chen, Chang Liu, and Dawn Song.
\newblock {Delving into Transferable Adversarial Examples and Black-box
  Attacks}.
\newblock \emph{ICLR (Poster)}, 2017.

\bibitem[Madry et~al.(2018)Madry, Makelov, Schmidt, Tsipras, and
  Vladu]{madry2018towards}
Aleksander Madry, Aleksandar Makelov, Ludwig Schmidt, Dimitris Tsipras, and
  Adrian Vladu.
\newblock Towards deep learning models resistant to adversarial attacks.
\newblock \emph{International Conference on Learn- ing Representations (ICLR)},
  2018.

\bibitem[Meng and Chen(2017)]{meng2017magnet}
Dongyu Meng and Hao Chen.
\newblock {Magnet: a two-pronged defense against adversarial examples}.
\newblock In \emph{CSS}, pages 135--147. ACM, 2017.

\bibitem[Metzen et~al.(2017)Metzen, Genewein, Fischer, and
  Bischoff]{MetzenGeneweinFischerEtAl2017}
Jan~Hendrik Metzen, Tim Genewein, Volker Fischer, and Bastian Bischoff.
\newblock {On Detecting Adversarial Perturbations}.
\newblock \emph{CoRR}, abs/1702.04267, 2017.
\newblock URL \url{http://arxiv.org/abs/1702.04267}.

\bibitem[Papernot et~al.(2016{\natexlab{a}})Papernot, McDaniel, Wu, Jha, and
  Swami]{papernot2016distillation}
Nicolas Papernot, Patrick McDaniel, Xi~Wu, Somesh Jha, and Ananthram Swami.
\newblock {Distillation as a defense to adversarial perturbations against deep
  neural networks}.
\newblock In \emph{2016 IEEE Symposium on Security and Privacy (SP)}, pages
  582--597. IEEE, 2016{\natexlab{a}}.

\bibitem[Papernot et~al.(2016{\natexlab{b}})Papernot, McDaniel, and
  Goodfellow]{PapernotMG16}
Nicolas Papernot, Patrick~D. McDaniel, and Ian~J. Goodfellow.
\newblock {Transferability in Machine Learning: from Phenomena to Black-Box
  Attacks using Adversarial Samples}.
\newblock \emph{CoRR}, 2016{\natexlab{b}}.

\bibitem[Papernot et~al.(2017)Papernot, McDaniel, Goodfellow, Jha, Celik, and
  Swami]{PapernotMcDanielGoodfellowEtAl2017}
Nicolas Papernot, Patrick~D. McDaniel, Ian~J. Goodfellow, Somesh Jha, Z.~Berkay
  Celik, and Ananthram Swami.
\newblock {Practical Black-Box Attacks against Machine Learning}.
\newblock In \emph{ACM AsiaCCS 2017}, pages 506--519, 2017.

\bibitem[Ren et~al.(2015)Ren, He, Girshick, and Sun]{RenHeGirshickEtAl2015}
Shaoqing Ren, Kaiming He, Ross~B. Girshick, and Jian Sun.
\newblock {Faster {R-CNN:} Towards Real-Time Object Detection with Region
  Proposal Networks}.
\newblock In \emph{NIPS}, pages 91--99, 2015.

\bibitem[Russakovsky et~al.(2015)Russakovsky, Deng, Su, Krause, Satheesh, Ma,
  Huang, Karpathy, Khosla, Bernstein, Berg, and
  Fei-Fei]{RussakovskyDengSuEtAl2015}
Olga Russakovsky, Jia Deng, Hao Su, Jonathan Krause, Sanjeev Satheesh, Sean Ma,
  Zhiheng Huang, Andrej Karpathy, Aditya Khosla, Michael Bernstein,
  Alexander~C. Berg, and Li~Fei-Fei.
\newblock {ImageNet Large Scale Visual Recognition Challenge}.
\newblock \emph{International Journal of Computer Vision (IJCV)}, 115\penalty0
  (3):\penalty0 211--252, 2015.

\bibitem[Shelhamer et~al.(2017)Shelhamer, Long, and
  Darrell]{ShelhamerLongDarrell2017}
Evan Shelhamer, Jonathan Long, and Trevor Darrell.
\newblock {Fully Convolutional Networks for Semantic Segmentation}.
\newblock \emph{IEEE Trans. Pattern Anal. Mach. Intell.}, 39\penalty0
  (4):\penalty0 640--651, April 2017.
\newblock ISSN 0162-8828.

\bibitem[Simonyan and Zisserman(2014)]{simonyan2014very}
Karen Simonyan and Andrew Zisserman.
\newblock Very deep convolutional networks for large-scale image recognition.
\newblock \emph{arXiv preprint arXiv:1409.1556}, 2014.

\bibitem[Simonyan and Zisserman(2015)]{SimonyanZisserman2015}
Karen Simonyan and Andrew Zisserman.
\newblock {Very Deep Convolutional Networks for Large-Scale Image Recognition}.
\newblock \emph{ICLR}, 2015.

\bibitem[Srisakaokul et~al.(2018)Srisakaokul, Zhong, Zhang, Yang, and
  Xie]{srisakaokul2018muldef}
Siwakorn Srisakaokul, Zexuan Zhong, Yuhao Zhang, Wei Yang, and Tao Xie.
\newblock {Muldef: Multi-model-based defense against adversarial examples for
  neural networks}.
\newblock \emph{arXiv preprint arXiv:1809.00065}, 2018.

\bibitem[Szegedy et~al.(2013)Szegedy, Zaremba, Sutskever, Bruna, Erhan,
  Goodfellow, and Fergus]{SzegedyZarembaSutskeverEtAl2013}
Christian Szegedy, Wojciech Zaremba, Ilya Sutskever, Joan Bruna, Dumitru Erhan,
  Ian~J. Goodfellow, and Rob Fergus.
\newblock {Intriguing properties of neural networks}.
\newblock \emph{CoRR}, abs/1312.6199, 2013.
\newblock URL \url{http://arxiv.org/abs/1312.6199}.

\bibitem[Szegedy et~al.(2014)Szegedy, Zaremba, Sutskever, Bruna, Erhan,
  Goodfellow, and Fergus]{Szegedy2014}
Christian Szegedy, Wojciech Zaremba, Ilya Sutskever, Joan Bruna, Dumitru Erhan,
  Ian~J. Goodfellow, and Rob Fergus.
\newblock Intriguing properties of neural networks.
\newblock In \emph{ICLR}, 2014.

\bibitem[Tram{\`{e}}r et~al.(2017)Tram{\`{e}}r, Kurakin, Papernot, Boneh, and
  McDaniel]{TramerKurakinPapernotEtAl2017}
Florian Tram{\`{e}}r, Alexey Kurakin, Nicolas Papernot, Dan Boneh, and
  Patrick~D. McDaniel.
\newblock {Ensemble Adversarial Training: Attacks and Defenses}.
\newblock \emph{CoRR}, abs/1705.07204, 2017.
\newblock URL \url{http://arxiv.org/abs/1705.07204}.

\bibitem[Tramer et~al.(2020)Tramer, Carlini, Brendel, and
  Madry]{tramer2020adaptive}
Florian Tramer, Nicholas Carlini, Wieland Brendel, and Aleksander Madry.
\newblock On adaptive attacks to adversarial example defenses.
\newblock \emph{arXiv preprint arXiv:2002.08347}, 2020.

\bibitem[Wang et~al.(2017)Wang, Zhang, Xie, Zhou, Premachandran, Zhu, Xie, and
  Yuille]{WangZhangXieEtAl2017}
Jianyu Wang, Zhishuai Zhang, Cihang Xie, Yuyin Zhou, Vittal Premachandran, Jun
  Zhu, Lingxi Xie, and Alan~L. Yuille.
\newblock {Visual Concepts and Compositional Voting}.
\newblock \emph{CoRR}, 2017.

\bibitem[Xiao et~al.(2017)Xiao, Rasul, and Vollgraf]{Han2017FashionMNIST}
Han Xiao, Kashif Rasul, and Roland Vollgraf.
\newblock Fashion-mnist: a novel image dataset for benchmarking machine
  learning algorithms.
\newblock 2017.
\newblock URL \url{http://arxiv.org/abs/1708.07747}.

\bibitem[Xie et~al.(2018)Xie, Wang, Zhang, Ren, and
  Yuille]{XieWangZhangEtAl2017}
Cihang Xie, Jianyu Wang, Zhishuai Zhang, Zhou Ren, and Alan Yuille.
\newblock {Mitigating adversarial effects through randomization}.
\newblock \emph{ICLR (Poster)}, 2018.

\bibitem[Yuan et~al.(2017)Yuan, He, Zhu, Bhat, and Li]{YuanHeZhuEtAl2017}
Xiaoyong Yuan, Pan He, Qile Zhu, Rajendra~Rana Bhat, and Xiaolin Li.
\newblock {Adversarial Examples: Attacks and Defenses for Deep Learning}.
\newblock \emph{CoRR}, abs/1712.07107, 2017.
\newblock URL \url{http://arxiv.org/abs/1712.07107}.

\end{thebibliography}
\bibliographystyle{plainnat}
\clearpage
% \normal
\appendix
\newpage

  \vbox{%
    \hsize\textwidth
    \linewidth\hsize
    \vskip 0.1in
  \hrule height 4pt
  \vskip 0.25in
  \vskip -5.5pt%
  \centering
    {\LARGE\bf{Buffer Zone based Defense against Adversarial Examples in Image Classification \\
    Supplementary Material} \par}
      \vskip 0.29in
  \vskip -5.5pt
  \hrule height 1pt
  \vskip 0.09in%
    
  \vskip 0.2in

  }

\appendix
\section{Adversarial model}
\label{sec:apdx:A1}

The strength of an attack is relative to the considered adversarial model. In adversarial Machine Learning (ML) the assumed capabilities of an attacker are a mix of:
\begin{description}
    \item[Having knowledge of the parameters and architecture of the defense network/classifier.] The architecture and methodology of how the defense network is trained is about how the defense operates and its underlying philosophy. In cryptography this is similar to the actual design of a security primitive and this is never assumed to be hidden as this would lead to the undesirable practice of security by obscurity. The secret key of the security primitive is kept private; the method of how the secrect key is generated using a probabilistic algorithm is public. Similarly, the parameters of a defense network can be considered to be private while the method of their training is public. If the adversarial model does not assume the parameters to be private, then all is public and we call this a {\bf white-box setting}. If the parameters are considered to be private and not given to the attacker (for `free'), then we call this a {\bf black-box setting}.
    \item[Having access to the defense classifier a.k.a. target model.] If the parameters are kept private, then the next best adversarial capability is having access to the target model as a black-box. The idea is that the adversary can adaptively query the target model with own input images for which class labels are being returned. Here, we have two flavors: only a class label is returned, or more information is given in the form of a score vector which entries represent confidences scores of each of the classes (the returned class label is the one with maximal confidence score). 
    
    In the white-box setting where all parameters are known, the attacker can reproduce the target model and access the target model as a black-box. Confusing in adversarial ML is that {\bf white-box attacks} are the ones that `only' use the parameters in the white-box setting to learn gradients of the target model which are used to produce adversarial examples -- these attacks do not consider/use black-box access. This means that white-box defenses are not necessarily analysed against {\bf black-box attacks} where the adversary only has black-box access to the target model with possibly the added capability described below.
    \item[Having access to (part of the) training data.] The training data which is used to train the parameters of the defense network is often public knowledge. Knowing the methodology of how the defense network is trained, an adversary can apply the same methodology to train its own synthetic defense network -- and this can be used to find adversarial examples. The synthetic network will not be exactly the same as the defense network since training is done by randomized (often SGD-like) algorithms where training data is used in random order. This means that knowledge of the training data is less informative than knowledge of the parameters as in the white-box setting.
\end{description}
The white-box setting describes the capabilities of the strongest adversary, while the black-box setting describes a weaker adversary who cannot exactly reproduce the target model (and can only estimate the target model by training a synthetic network). White-box attacks on the other hand restrict the adversary in that only oracle access to the gradient of $f$ is allowed. Black-box attacks only allow oracle access to the target model itself and oracle access to training data. In this sense white-box attacks in adversarial ML literature exclude access to the above black-box oracles. This means that even though a white-box defense may be able to resist a white-box attack, it can still be vulnerable to a black-box attack. Vice versa, even though a white-box defense may be broken under a white-box attack, it may still survive in the black-box setting.

Taking BUZz as an example, we may mathematically formalize the black-box adversary\footnote{Similarly a white-box adversary with only oracle access to gradient information can be modeled.}  (as is done in crypto/security literature) as an adversarial algorithm ${\cal A}_T$ which has access to 
\begin{itemize}
    \item a random oracle representing $l\leftarrow \mbox{BUZz}_{\theta}(x)$ where parameters $\theta = (c_j, i_j, w_j)_{j=1}^m$, input $x$ is an image, and $l$ is the outputted label of the target classifier BUZz (we denote the collected $(x,l)$ by set $\mathcal{X}_1$), and
    \item a random oracle $\xi_D$ which outputs at most $D$ times `training data' according to the distribution from which the training data is taken from (we denote the collected $(x,l)$ by set $\mathcal{X}_0$ which represents the 'original training data'); by abuse of notation $\xi$ denotes the distribution itself. 
\end{itemize}
Subscript $T$ denotes the allowed number of  computation steps plus oracle accesses to $\mbox{BUZz}_{\theta}$ and $\xi_D$. In our experiments we test the most powerful existing black-box attacks and do not mention $T$; here, $T$ just means the amount of steps allowed by existing practical attack methodologies. If an attacker with unlimited access ($T=\infty$) to $\mbox{BUZz}_{\theta}$ can scan region boundaries, then this achieves optimal success rates.

Subscript $D$ in $\xi_D$ indicates the number of training data an attacker is allowed to use for the attack. $D$ represents an important metric in machine learning as the amount of training data cannot be assumed infinite (with respect to the application there is a concrete limit to how many training data is available; collecting samples from $\xi$ is not straightforward, e.g., making a true new picture/image of a plane takes effort). How
strong we make e.g. Papernot's black-box attack is based on how much training data we give it: In their paper $D=150$  is used for MNIST in order to train a synthetic network while in our experiments we use $D=50$K which is the {\em entire} original training data set of CIFAR-10 (and leaves $10$K test data). %This makes the attack stronger and 
Since the attack uses the synthetic network in a (targeted or untargeted) white-box attack (with small enough $\epsilon$) to generate an adversarial example, the probability of successfully changing the label depends on how similar the synthetic network classifies data compared to the target model with defense -- it depends on the tranferability between the synthetic and defense classifiers and transferability improves for larger $D$.

The aim of the adversary is to produce a perturbation $\eta \leftarrow {\cal A}_T^{\xi_D, \mbox{BUZz}_{\theta}}(x)$ (just based on the oracle accesses described above indicated by superscripts and based on input $x$) which is visually imperceptible, i.e. $\|\eta\|\leq \epsilon$, and for which $x'=x+\eta$ gives a different label (and is not classified by the adversarial label $\bot$): 
%The miss classification rate in the presence of an attacker
The attacker's success rate for 
%with respect to
untargeted black-box attacks is defined as the probability
$$\alpha = Pr_{x\leftarrow \xi}( \eta \leftarrow {\cal A}_T^{\xi_D, \mbox{BUZz}_{\theta}}(x), \ \|\eta\|\leq \epsilon, \ \mbox{BUZz}_{\theta}(x+\eta) \notin \{\bot,
lab(x)\} \ | \ \mbox{BUZz}_{\theta}(x)=lab(x) ),$$
%\mbox{BUZz}_{\theta}(x)\} )$$
where $lab(x)$  represents the correct label according to the human eye.

For targeted black-box attacks we replace $\mbox{BUZz}_{\theta}(x+\eta) \notin \{\bot, lab(x)\}$ 
%\mbox{BUZz}_{\theta}(x)\}$ 
by $\mbox{BUZz}_{\theta}(x+\eta)=l'$, 
%replace $\mbox{BUZz}_{\theta}(x) \neq \bot$ by $\mbox{BUZz}_{\theta}(x) \notin \{\bot, l'\}$, 
and take the probability over both $x\leftarrow \xi$ and $l'\leftarrow \{1..k\}\setminus \{lab(x)\}$. In the above notations we do not explicitly state that the adversary also has knowledge of the distribution from which $\theta$ is taken, i.e., the adversary knows the philosophy behind our defense together with what type of  image transformations are being used and knows the architecture in terms of number of nodes and connections at each layer of the CNN networks and how they are trained.

The above formalism helps in making the adversarial model in terms of adversarial capabilities precise. We will not explicitly use the formalism as we cannot prove statements about general classes of adversarial algorithms ${\cal A}$ (our defense does not allow `standard' reduction proofs to some hard computational problem as is done in crypto).

From a cryptographer's perspective we want the above probability $\alpha$ (i.e., the attacker's success rate) to be as small as possible. Notice that the smallest possible $\alpha=0$ decreases the miss classification rate to $1-(p-\gamma)\geq 1-p$ (see Section \ref{sec:metric}, the miss classification rate as a function of $\alpha$ is equal to  $1-(p-\gamma)(1-\alpha)$). Since in practical ML the miss classification rate $1-p$ for vanilla schemes (without defense) is strictly larger than $0$, it makes no sense to require (as we often do in crypto) an asymptotic guarantee like demanding $\alpha$ to be negligible in some security parameter $\lambda$ where images $x$ are $poly(\lambda)$-sized. In stead we measure $\alpha$ in percentages and a couple percent is considered very small.

%Preferably we want the advantage $|a'-a|$ with 
%$$
%a = Pr_{x\leftarrow \xi}(  \mbox{BUZz}_{\theta}(x) \neq label(x) )
%$$
%to be small, i.e., misclassification 
%Notice that $a=1-p$

%even negligible in some security parameter $\lambda$ where image $x$ is $poly(\lambda)$-sized. First, in ML we have concrete data sets with images of certain fixed sizes for which we want to design defense strategies against adversarial ML. So, such an asymptotic goal makes no sense. Second, it turns out that it is very difficult to obtain a defense strategy that can make the attacker's success rate very small, say 0.1\%, without sacrificing the clean accuracy all together. In this paper we have been able to reduce the attacker's success rate down to about $<10$\% while only reducing the clean accuracy by 8-18\%.

\section{Related work: comparison to known Defenses}
%\label{sec:defense}
\label{app:rel}

\subsection{White-box attacks} \label{app-white}

As explained and motivated in the introduction, we restrict ourselves to the black-box setting where the parameters of our defense are kept secret.  Hence, this disallows direct white-box attacks and zeroth order optimization based black-box attacks. However, it is important to note that once a synthetic model has been trained, any white-box attack can be run on the synthetic model to create an adversarial example. The adversary can then check if this example fools the defense. 

Essentially any white-box attack can be run on the synthetic model to try to exploit the transferability between classifiers~\cite{PapernotMG16}. We briefly introduce the following commonly used white-box attacks in the literature. 

We briefly introduce the following  commonly used white-box attacks in  literature. 

%\begin{itemize}
\noindent
%\item
\textbf{Fast Gradient Sign Method (FGSM)~\cite{GoodfellowShlensSzegedy2014}.} $x' = x' + \epsilon  \times sign(\nabla_x L(x,l;\theta)$ where $L$ is a loss function (e.g, cross entropy) of model $f$. 

\noindent
%\item
\textbf{Basic Iterative Methods (BIM)~\cite{kurakin2017adversarial}.}
$x'_i = \text{clip}_{x,\epsilon}(x'_{i-1} + \frac{\epsilon}{r}  \times sign(\nabla_{x'_{i-1}} L(x'_{i-1},l;\theta))$ where $x'_0=x$, $r$ is the number of iterations, \text{clip} is a clipping operation.

\noindent
%\item
\textbf{Momentum Iterative Methods (MIM)~\cite{dong2018boosting}.}
This is a variant of BIM using momentum trick to create the gradient $g_i$, i.e., $x'_i = \text{clip}_{x,\epsilon}(x'_{i-1} + \frac{\epsilon}{r}  \times sign(g_i))$. 

\noindent
%\item
\textbf{Projected Gradient Descent (PGD)~\cite{madry2018towards}.}
This is also a variant of BIM where the clipping operation is replaced by a projection operation. 

\noindent
%\item 
\textbf{Carlini and Wagner attack  (C$\&$W)\cite{carlini2017adversarial}.}
We define $x'(\omega)=\frac{1}{2}(\tanh{\omega}+1)$ and $g(x)=\max( \max(s_i: i\neq l) -s_i, -\kappa)$ where $f(x)=(s_1,s_2,\ldots)$ is the score vector of input $x$ of classifier $f$ and $\kappa$ controls the confidence on the adversarial examples. The adversary builds the following objective function for finding the adversarial noise.
$$
\min_{\omega}\|x'(\omega) - x \|^2_2 + c f(x'(\omega)), 
$$
where $c$ is a constant chosen by a modified binary search. 

\noindent
%\item 
\textbf{Elastic Net Attack (EAD)~\cite{chen2018ead}.} This is the variant of C$\&$W attack with the following objective function. 
% Removed empty line
$$
\min_{\omega}\|x'(\omega) - x \|^2_2 + \beta\|x'(\omega) - x \|_1 + c f(x'(\omega)).  
$$

%\end{itemize}

\subsection{Black-box attacks.} \label{app-black}
%A defense seems to be vulnerable if the adversary completely or partially knows about the defense, i.e., using the gradient $\nabla f$ of classifier $f$ (white-box attack) or using 
Black-box attacks use non-gradient information of classifier $F$ such as (part of) the original training data set $\mathcal{X}_0$~\cite{PapernotMG16} and/or a set $\mathcal{X}_1$ of addaptively chosen queries to $F$ (i.e.,  
%$\{(x,f(x)) : x\in \mathcal{X}_1\}$ or 
$\{(x,l=F(f(x))) : x\in \mathcal{X}_1 \}$)~\cite{PapernotMcDanielGoodfellowEtAl2017} -- queries in $\mathcal{X}_1$ are not in the training data set $\mathcal{X}_0$. %Indeed, many attacks based on non-gradient information, black-box attack, 
These type of attacks exploit the \textit{transferability} property of adversarial examples \cite{PapernotMG16,LiuChenLiuEtAl2016}: %The transferability property of adversarial examples means the 
Based on information $\mathcal{X}_0$ and $\mathcal{X}_1$ the adversary trains its own copy of the proposed defense. This is called the adversarial synthetic network/model and is used to  create adversarial examples for the target model. %The adversarial examples are submitted to the considered defense's model. 
\cite{LiuChenLiuEtAl2016} shows that the transferability property of adversarial examples between different models which have the same topology/architecture and are trained over the same dataset is very high, i.e., nearly $100\%$ for  ImageNet \cite{RussakovskyDengSuEtAl2015}. This explains why the adversarial examples generated for the synthetic network can often be successful adversarial examples for the defense network.

Black-box attacks can be partitioned into the following categories:

\noindent
\textbf{Pure black-box attack}~ \cite{Szegedy2014,PapernotMG16,AthalyeCarliniWagner2018,LiuChenLiuEtAl2016}. The adversary is \textit{only} given  knowledge of a training data set $\mathcal{X}_0$. % (i.e., the distribution from which $\mathcal{X}$ is sampled). %or $\mathcal{X}$. 
    Based on this information, the adversary builds his own classifier $g$ %(see Algorithm~\ref{alg:pure_synthetic}) 
    which is used to produce adversarial examples using an existing white-box attack methodology. These adversarial examples of $g$ may also be the adversarial examples of $f$ due to the transferability property between $f$ and $g$.
    
\noindent 
\textbf{Oracle based black-box attack}~\cite{PapernotMcDanielGoodfellowEtAl2017}. The adversary is allowed to adaptively query  target classifier $F$, which gives information $\mathcal{X}_1$.
    %i.e., the $\{x,f(x)\}$ or $\{x,F(f(x))\}$. \textit{The queries $x$s are not in the training dataset of target model $f$~\cite{PapernotMcDanielGoodfellowEtAl2017}}.
    Based on this information, the adversary builds his own classifier $G$ which is used to produce adversarial examples
    using an existing white-box attack methodology.
    %later (see Algorithm~\ref{alg:papernot_synthetic}). 
    Again, the generated adversarial examples for $G$ may also be able fool  classifier $F$ due to the transferability property between $F$ and $G$. Compared to the native (pure) black box attack, this attack is supposed to be more efficient because $G$ is intentionally trained to be similar to $F$. Hence, the transferability between $F$ and $G$ may be significantly higher.   
    
\noindent
\textbf{Zeroth order optimization based black-box attacks or Score Value Vector based black-box attacks}~\cite{ChenPinYu2017}. The adversary does not build any assistant classifier $g$ as done in the previous black-box attacks. Instead, the adversary  adaptively queries $\{x,f(x),F(f(x))\}$ to approximate the gradient $\nabla f$ %(which cannot be directly computed due to lack of knowledge on the structure of $f$) 
    based on a derivative-free optimization approach. Using the approximated $\nabla f$, the adversary can build adversarial examples by directly working with the classifier $f$. Another attack in this line is called SimBA (Simple Black Box Attack)~\cite{guo2019simple}. This attack also requires the score vector $f(x)$ to mount the attack.   
    
\noindent
\textbf{Decision based black-box attack~\cite{Chen2019Hop}}. As shown in the literature, the adversary is still able to mount a black box attack if he is only allowed to access the decision value, i.e., $F(f(x))$. The main idea of the attack is to try to find the boundaries between the class regions using a binary searching methodology and gradient approximation for the points located on the boundaries. This type of attacks are called  Boundary Attacks. As mentioned in~\cite{Chen2019Hop} Boundary Attacks are not efficient as a pure black box attack.

In this paper, we analyze  a {\bf mixed black-box attack} %algorithm 
where the synthetic network $g$ (which outputs the score vector for the full classifier $G$) is built based on the training data set $\mathcal{X}_0$ of the target model $F$ and is based on addaptively chosen queries $\mathcal{X}_1$. % Papernot's method (see Algorithm~\ref{alg:our_papernot_synthetic} for more detail). 
 Our mixed black-box attack is more powerful than both the pure black-box attack and oracle based black-box attack. We further confirm this fact through our experiments in Supplemental Material~\ref{expana} 
 (see Tables~\ref{table:pureBB_Fashion_CIFAR},
 \ref{table:mixBB_Fashion_all_BUZz} and~\ref{tab:all_attack_all_defenses_CIFAR10}).

\subsection{Defenses against white-box and black-box attacks} \label{literature}

\noindent
{\bf White-Box defenses.}
%In this section we consider 
White-box defenses are any defense with an adversarial model that allows the adversary oracle access to the gradient of the target model. These defenses include~\cite{papernot2016distillation,KurakinGoodfellowBengio2016,TramerKurakinPapernotEtAl2017,CaoGong2017,MetzenGeneweinFischerEtAl2017,FeinmanCurtinShintreEtAl2017,XieWangZhangEtAl2017,meng2017magnet,srisakaokul2018muldef} to name a few. A complete list is given in \cite{AthalyeCarliniWagner2018,carlini2017adversarial} except for the unpublished work \cite{srisakaokul2018muldef} which appeared later.
%\textbf{Observation 1.} 
So far, any defense with public weights and architecture turns out to be vulnerable to FGSM, IFGSM, or Carlini type attacks~\cite{carlini2017adversarial,CarliniWagner2017,AthalyeCarliniWagner2018,LiuChenLiuEtAl2016}; we will argue the vulnerability of \cite{srisakaokul2018muldef} below.

In order to implement a white-box attack
\cite{YuanHeZhuEtAl2017} constructs perturbation $\eta$  with the help of gradient $\nabla f(\cdot)$: for example, $\eta=\epsilon \times sign(\nabla_x L(x,l;\theta)$ in the Fast Gradient Sign Method (FGSM) by~ \cite{GoodfellowShlensSzegedy2014}, where $\theta$ represents the parameters of $f$, $L$ is a loss function (e.g, cross entropy) of model $f$. ($\epsilon$ can be thought of as relating to the maximum amount of noise which is not visually perceptible.) In
Appendix~\ref{app-white},
%Section~\ref{sec:background}, 
the short description of some widely known white-box attacks is provided. To defend against adversarial examples, many methodologies have been proposed and they all employ the same strategy, i.e., \textit{gradient masking} \cite{PapernotMcDanielGoodfellowEtAl2017} respectively \textit{obfuscated gradient} \cite{AthalyeCarliniWagner2018}. As pointed out in~\cite{AthalyeCarliniWagner2018}, there are three main
methods for realizing this strategy: %strategies: 
\textit{shattered gradients}, \textit{stochastic gradients} and \textit{exploding $\&$ vanishing gradients}. 
In \cite{AthalyeCarliniWagner2018}, the authors propose three different types of attacks: 
\begin{enumerate}
\item \textbf{Backward Pass Differentiable Approximation (BPDA)}. The attack is applied for protected network $f(t(x))$ where $t(x)$ is not differentiable and $t(x)\approx x$. The adversary will replace $t(x)$ in the backward phase for computing the gradient by $x$ and thus, he can compute the approximated gradient $\nabla_x f(t(x))|_{x=\hat{x}} \approx \nabla_x f(x)|_{x=t(\hat{x}}$. 
\item \textbf{Expectation over Transformation (EOT)}. In this case, the adversary computes the gradient of $\mathbb{E}_{t \sim T} f(t(x))$ where $t(x)$ is a random transformation and $t$ is sampled from a distribution $T$. The gradient can be computed as $\nabla \mathbb{E}_{t \sim T} f(t(x)) = \mathbb{E}_{t \sim T} \nabla f(t(x))$. 

\item \textbf{Reparameterization.} The protected network $f(t(x))$ has $t(x)$ which performs some optimization loop to transform the input $x$ to a new input $\hat{x}$. This step will make the gradient exploding or vanishing, i.e., the adversary cannot compute the gradient. To cope with this defense, \cite{AthalyeCarliniWagner2018} proposes to make a change-of-variable $x=h(z)$ for some $h(\cdot)$ such that $t(h(z))=h(z)$ for all $z$ but $h(\cdot)$ is differentiable.  
\end{enumerate}

In  literature, many white-box defenses have shown a predictable cat and mouse type of security game. In this repeated chain of events, the defender creates a network defense and the attacker comes up with a new type of attack that breaks the defense. The defender then creates a new defense which the attacker again breaks. While this occurs frequently in security, a simple example of this occurring in the field of adversarial machine learning is the FGSM attack breaking standard CNNs, the distillation defense mitigating FGSM, and the distillation defense’s subsequent break by Carlini~\cite{papernot2016distillation,CarliniWagner2017}. Alternatively, in an even worse case, the defense can be immediately broken without the need for new attack strategies. In adversarial machine learning an example of this is %Meng and Chen et al. 
the autoencoder defense of \cite{meng2017magnet} which is vulnerable to the attack in %and Carlini et al.~
\cite{CarliniWagner2017}.
%attack. 
%
From our analysis of the previous literature it is clear that a secure pure white-box defense is extremely challenging to design.

\noindent
{\bf Black-box defenses based on a single network.}
We discuss how the white-box defenses of~\cite{papernot2016distillation,KurakinGoodfellowBengio2016,TramerKurakinPapernotEtAl2017,CaoGong2017,MetzenGeneweinFischerEtAl2017,FeinmanCurtinShintreEtAl2017,XieWangZhangEtAl2017,meng2017magnet,GuoRanaCisseEtAl2017,srisakaokul2018muldef} are vulnerable in a black-box setting. %are 

As shown in~\cite{PapernotMcDanielGoodfellowEtAl2017}, the adversary can build a synthetic network $g$ which simulates the target vanilla network. This can be used to produce high transferability adversarial examples (that transfer to the target model with significant success). 
%The well-known explanation for this problem is 
Boundary alignment is the well-known explanation, see~\cite{PapernotMG16}. 
%Now we consider the single network with some defense operations. 

\cite{papernot2016distillation} proposes a single network defense with a better adversarial robustness property based on distillation: First, given a training data set, a network is built and trained. After this, the softmax output (i.e., score vector) of the network is used to train another network with the original training data set. This process is called 'distillation' and the distilled network is argued to have better robustness against white-box attacks. However, \cite{CarliniWagner2016} showed a white-box attack against this defense. Moreover,  \cite{PapernotMcDanielGoodfellowEtAl2017} showed that for the MNIST dataset, the success rate of Papernot's black-box attack (untargeted) is at least $70\%$. 

%the distilled network is also vulnerable to Papernot's black-box attack.  

In~\cite{KurakinGoodfellowBengio2016}, the authors discuss how to train the network for ImageNet with adversarial examples to make it robust against adversarial machine learning. During each epoch in the training process, adversarial examples are generated and again used in the training process. According to %the result reported in 
Table 4 in~\cite{KurakinGoodfellowBengio2016}, the success rate of (untargeted) pure black box attack on ImageNet using FGSM is high  $\geq 50\%$. The authors in \cite{TramerKurakinPapernotEtAl2017} also claim that the adversarial training in~\cite{KurakinGoodfellowBengio2016} may not be useful.

\cite{TramerKurakinPapernotEtAl2017} proposes another type of adversarial training method. The adversarial examples are generated by doing attacks on different networks with different attack methods. After this the designer trains the new network with the generated adversarial examples. The authors argued that this adversarial training can make the adversarially trained network more robust against (pure) black-box attacks because it is trained with adversarial examples from different sources (i.e., pre-trained networks). In other words, the network is supposed to have better robustness against black-box attack generalization across models. As shown in~\cite{AthalyeEngstromIlyasEtAl2018}, the adversarially trained network is vulnerable to white-box attack. Regarding pure black-box attack, as reported in Table 4 in~\cite{TramerKurakinPapernotEtAl2017}, the success rate of (untargeted) pure black box attack on ImageNet using FGSM -- the best known black box attack that has been executed on this defense -- is $\geq 27\%$. We verify the efficiency of this approach for CIFAR-10 in this paper. We do the adversarial retraining using data from 5 other networks to build a 6th network. The 5 other networks are from the Mul-Def paper~\cite{srisakaokul2018muldef} (we also rigorously discuss this paper next few paragraphs). Network 0, is a vanilla ResNet. Network 1 is a ResNet with 30$\%$ adv training from network 0. Network 2 is a ResNet with 30$\%$ adv training (15$\%$ from network 0, 15$\%$ from network 1). Etc. until we get to Network 5. After training we run the full Papernot attack on Network 5. The full result we can find in Table~\ref{tab:all_attack_all_defenses_CIFAR10}, i.e., Adv Retrained 1-Net. The defense has clean prediction accuracy of 86$\%$ and the best attack on the defense is untargeted PGD with success rate of $55\%$.

\cite{CaoGong2017} proposes a white-box defense based on the following trick: for a given sample $x$, the defense collects many samples $x'_1,\cdots,x'_n$ in a small hypercube centered at $x$. Then, the outputted class label is the one which gains the  majority vote among $F(f(x'_1)),\cdots,F(f(x'_n))$ where $F$ is the output function of network $f$. 
As discussed in~\cite{cohen2019certified}, this is a very first implementation of a class of defense called "Randomized Smoothing" technique. In Table~\ref{tab:all_attack_all_defenses_CIFAR10}, we perform mixed black box attacks on this defense and see that the defense is vulnerable to mixed black box attacks, i.e., the defense has clean prediction accuracy of 92$\%$ and the best attack on the defense is untargeted MIM with success rate of $61\%$. 

In~\cite{MetzenGeneweinFischerEtAl2017}, the authors constructed a defense of a single network $f$ with an additional `detector' network $g$. The `detector network' is built based on the training data set of the main network $f$ together with adversarial examples generated for  training data samples. The detector network is used to distinguish clean samples from adversarial samples. The authors in~\cite{carlini2017adversarial} showed  white-box and black-box attacks on this defense. The success rate of untargeted pure black-box attack on MNIST using the C$\&$W attack by~\cite{CarliniWagner2016} is at least $84\%$. 

In~\cite{FeinmanCurtinShintreEtAl2017}, the authors built a detector to distinguish adversarial examples from clean examples using Bayesian uncertainty estimate or Kernel Density Estimator. The key idea is that since the adversarial and clean examples do not belong to the same manifolds, the defender can build a detector. \cite{carlini2017adversarial} showed a white-box attack on this defense  %. Especially, the authors 
and clearly claim that the defense does not work if the dataset is CIFAR-10 for both white-box attack and black-box attack, i.e., the success rate of untargeted pure attack seems at least $80\%$ based on their explanation. % of the authors.  

\cite{XieWangZhangEtAl2017} has a single network and uniformly selects an image transformation from an a-priori fixed set of a small number of image transformations 
%a fixed randomized image transformation with a single network
%As shown in~\cite{XieWangZhangEtAl2017}, the designer applied the randomization operation to a single network 
to defeat white-box attacks. In the white-box setting \cite{AthalyeCarliniWagner2018} shows that this defense does not work. But is this defense secure against black-box attacks? 
%Then answer is negative. Indeed, 
To maintain a sufficiently high clean accuracy, the random image transformations should not have high randomness. Hence, the boundaries of any single network/classifier and  the network/classifier with one of the random image transformations may be highly aligned. This implies that the adversarial examples created for any classifier %would be highly transferred 
will likely transfer to the network with randomization operations. This is confirmed by experiments reported in Table~\ref{tab:all_attack_all_defenses_CIFAR10}. We can see that the  defense accuracy (i.e., clean accuracy for the defense model) is $82\%$ and the attacker's success rate of the untargeted mixed  black-box attack (using MIM)
%using  best attack of Papenot's attack -- iFGSM-- 
is $73\%$. 

%The similar reason can be found to explain why 
Similarly, the defense proposed in~\cite{meng2017magnet} -- a defense with a single network and multiple different auto-encoders as image transformations  from which one is selected at random per query %used as random input transformations is proposed
--  is not secure against pure black-box attacks, i.e., the success rate of targeted pure black-box attack on the defense using C$\&$W attack for CIFAR-10 and MNIST is at least $99\%$, see  ~\cite{CarliniWagner2017}. % shows that this defense is broken using pure black-box attack with small noise.

In~\cite{GuoRanaCisseEtAl2017}, the designer selects a set of possible image transformations for a single network and keeps the selection of the chosen image transformation secret. The image transformation will distort the noise as explained in~\cite{GuoRanaCisseEtAl2017}. This is BUZz for a single protection layer (without multiple networks and threshold voting).
%all of them in secret to build a defense. 
%The image transformation will distort the noise as explained in~\cite{GuoRanaCisseEtAl2017}. 
However, %as explained in Section~\ref{sec:intro},
there is no buffer zones for any single network and thus, there exist many adversarial examples with small noises. %We will do black-box attack in Section~\ref{sec:experiment} to validate this claim. 
We have validated this claim for BUZz with a single protection layer because it is very close to the one in~\cite{GuoRanaCisseEtAl2017}. In Table~\ref{tab:all_attack_all_defenses_CIFAR10}, the best attack is the untargeted one with a success rate 
%of the best attack on the defense 
for CIFAR-10 equal to $56\%$ and a clean accuracy equal to $90\%$. 

\noindent
{\bf Black-box defenses based on multiple networks.}
In~\cite{LiuChenLiuEtAl2016}, the authors study the transferability between different networks which have different structures for the ImageNet dataset. The authors report that the transferability between the networks is small (claimed to be 'close to zero'). For this reason, it may be possible to have the BUZz defense where protected networks have different architectures.
%since all the networks seem satisfy two aforementioned defense properties? The answer is negative. 
However, experimentally we have built a defense with VGG16 and ResNet56 trained on CIFAR-10. As reported in Table~\ref{tab:all_attack_all_defenses_CIFAR10}, the 2-net BUZz with 2 ResNet56 architectures and image transformations and the 2-net defense with 1 VGG16 and 1 ResNet56 (Mixed Arch 2-Net) that has no image transformations has worse performance. It implies that having different architectures does not give us a significant advantage in a BUZz defense.

In unpublished work~\cite{srisakaokul2018muldef} the authors have proposed a defense against white-box attacks based on multiple networks with the \textit{same} architecture. The authors develop their defense based on  a retraining technique. First, the authors apply adversarial attacks on each network to generate  a set of adversarial examples. For example, for each network $j$ a white-box attack produces a set of adversarial examples $\mathcal{S}_j$. Next network $j$ will be retrained with the clean training data set together with some of the adversarial sets $\mathcal{S}_h$, $h\neq j$. The authors argue that all the networks cannot be fooled at the same time for a given adversarial example and this leads to a low(er) attacker's success rate. The final outputted class label  is the predicted label of one of the networks chosen at random among all networks; this gives high clean accuracy.
%which is randomly chosen among all networks. This allows the defense achieves Defense Property 2. 

%\textcolor{red}{
With respect to white-box attacks, the defense in~\cite{srisakaokul2018muldef} seems not secure: For verifying resistance against white-box attacks, \cite{srisakaokul2018muldef}, %to verify the resistance of the defense against the white-box attack, the authors 
only attacks each model separately instead of attacking all the models at the same time as is done in Ensemble Pattern Attacks in Section 4 in~\cite{XieWangZhangEtAl2017}. Hence, the authors should do the Ensemble Pattern Attack on their defense to have a completed claim on white-box resistance.
%}

%\textcolor{red}{
For testing resistance against Papernot's black-box attack, the authors only work with an initial set of 150 samples and 5 runs. This gives an augmented set of only $2^4\cdot 150 =2400$ samples used for building the synthetic model (compared to an augmented set of $2^5\cdot 50$K samples in our experiments). Hence, the performance of synthetic model $g$ is very poor and as a result a lower attacker's success rate. Nevertheless, even with a poor synthetic network (that is, very weak black-box adversary) the reported  success rate of the Papernot's attack with FGSM is still high, i.e., around $18\%/ 27\%$ for MNIST and CIFAR-10 (see Table 5 in~\cite{srisakaokul2018muldef}.   
%}
We performed experiments for our strong mixed black-box adversary and found an attacker's success rate of $63\%$ for the best attack on the best defense (Mul-Def 4 (MD4) Network in Table~\ref{tab:all_attack_all_defenses_CIFAR10}).

We summarize the attacker's success rate of the black-box attacks on the aforementioned defenses in
Table~\ref{table:success_rate_bb}. We can see that all defenses do not want to sacrifice clean accuracy $p$ -- generally at most a drop of about 5\%, 
see \cite{meng2017magnet}, with a larger drop of 13\% in \cite{XieWangZhangEtAl2017}. As a result of aiming at keeping the clean accuracy the same, the attacker's success rate $\alpha$ remains very high, typically $\alpha\geq 0.50$ for experiments with Fashion-MNIST, CIFAR-10 as in this paper.

These high attacker's success rates give a drop in the effective clean accuracy of $\delta=p - (p-\gamma)(1-\alpha)\geq p - p(1-0.5)= 0.5 p$; the $\delta$ values are listed in the table. 

For ImageNet, not studied in our paper, we have 
%Only the defenses for ImageNet lead to 
$\delta$ equal to 0.211 in \cite{TramerKurakinPapernotEtAl2017} and 0.396 in \cite{KurakinGoodfellowBengio2016}, albeit for the weaker pure black-box adversary (without access to the entire training data set). First, for the stronger adversary analysed in this paper we expect these $\delta$ values to become higher. 
Second, since 
 \cite{TramerKurakinPapernotEtAl2017}  has (compared to $\delta=0.247$ for BUZz) a  higher $\delta=0.29$ for the stronger adversary for CIFAR-10, we expect to also see this reflected in a smaller $\delta$ for ImageNet for BUZz compared to  \cite{TramerKurakinPapernotEtAl2017}  (after fine tuning image transformations).
We leave exact experiments for future work.
Third, even if BUZz would do similar when comparing $\delta$ to \cite{TramerKurakinPapernotEtAl2017} for ImageNet, BUZz would still be the better choice: This is because even though the $\delta$ values are the same, the attack success rate for BUZz is expected to be 7-10\%, much less than $~27\%$ of \cite{TramerKurakinPapernotEtAl2017} for the weaker adversary. This means that there is much less adversarial control in BUZz compared to \cite{TramerKurakinPapernotEtAl2017}. 
%(This has already been shown to be true for CIFAR-10 in Table  \ref{table:success_rate_bb}.)

\begin{table*}[th!]
\centering
\caption{Attacker's success rate of  black-box attacks for state-of-the-art defenses}
\scalebox{0.62}{
\begin{tabular}{|c|c|c|c|c|c|c|c|c|} 
 \hline
 Defense & Data set & Attack & Att. success rate $\alpha$ & Orig. Cl.Ac $p$& Cl.Ac $p_d$ & $\delta$ & $\gamma$\\ [0.5ex] 
 \hline\hline
 \cite{TramerKurakinPapernotEtAl2017} & ImageNet & Pure BB -FGSM & $\approx 27\%$ \cite{TramerKurakinPapernotEtAl2017} & $78\%$ (ImageNet)& $78\%$ (ImageNet)
& 0.211 & 0.0\\ \hline 
\cite{KurakinGoodfellowBengio2016} & ImageNet & Pure BB- FGSM & $\geq 50\%$~\cite{KurakinGoodfellowBengio2016} & $79\%$(ImageNet)& $78\%$ (ImageNet)
& 0.396 & 0.01\\ \hline \hline
%  BUZz & CIFAR-10 & Mixed BB - iFGSM & $6.9\%$ (this paper) & $88.35\%$ (CIFAR-10) & $68.32\%$ (CIFAR-10)
% & 0.247 \\ \hline
 \cite{TramerKurakinPapernotEtAl2017} & CIFAR-10 & Mixed BB -MIM & $55\%$ (this paper) & $93\%$ (CIFAR-10) & $86\%$ (CIFAR-10)
 & 0.54 & 0.07\\ \hline 
 
 %\textcolor{red}{ No defense} & CIFAR-10 & Mixed BB - iFGSM & $72\%$ (this paper) & $88\%$ (CIFAR-10) & $88\%$ (CIFAR-10)
% & \textcolor{red}{0.63} \\   \hline 
   \cite{GuoRanaCisseEtAl2017} & CIFAR-10 & Mixed BB - MIM & $56\%$ (this paper) & $93\%$ (CIFAR-10) & $90\%$ (CIFAR-10) 
 & 0.53 & 0.03\\ \hline
 
    \cite{CaoGong2017,cohen2019certified} & CIFAR-10 & Mixed BB - MIM & $61\%$ (this paper) & $93\%$ (CIFAR-10) & $92\%$ (CIFAR-10) 
 & 0.57 & 0.01 \\ \hline
 
   \cite{srisakaokul2018muldef} & CIFAR-10 & Mixed BB - MIM & $63\%$ (this paper) & $93\%$ (CIFAR-10) & $88\%$ (CIFAR-10)
 & 0.60 & 0.05\\   \hline

  \cite{FeinmanCurtinShintreEtAl2017} & CIFAR-10 & Pure BB - C$\&$W & $\geq 80\%$ ~\cite{carlini2017adversarial} & $83\%$ (CIFAR-10) & $83\%$ (CIFAR-10)
 & 0.661 & 0.0\\ \hline
 \cite{XieWangZhangEtAl2017} & CIFAR-10 & Mixed BB - MIM & $73\%$ (this paper) & $93\%$ (CIFAR-10) & 
 $82\%$ (CIFAR-10)
 % $0.59\%$ (CIFAR-10)
 & 0.72 & 0.11\\   \hline
\cite{meng2017magnet} & MNIST $\&$ CIFAR-10 & Pure BB - C$\&$W & $\geq 99\%$ ~\cite{CarliniWagner2017} &$91\%$ (CIFAR-10) & $87\%$ (CIFAR-10)
 & 0.897 & 0.04\\  \hline \hline
 \cite{papernot2016distillation} & MNIST & Oracle BB - FGSM & $70\%$ \cite{PapernotMcDanielGoodfellowEtAl2017} & $99\%$ (MNIST) & $98\%$ (MNIST)
 & 0.701 & 0.01\\ \hline 
\cite{MetzenGeneweinFischerEtAl2017} & MNIST & Pure BB - C$\&$W & $\geq 84\%$~\cite{CarliniWagner2017}& $92\%$ (MNIST) & $92\%$ (MNIST) & 0.769 & 0.0\\ \hline
\end{tabular}
}
% \caption{Attacker's success rate of  black-box attacks for state-of-the-art defenses}
\label{table:success_rate_bb}
\end{table*}

%\clearpage
\section{Pseudo algorithms: Black-box attack \& BUZz}
\label{app:pseudo_alg}

\noindent
{\bf Synthetic network.}
Algorithm~\ref{alg:papernot_synthetic}  depicts the construction of a synthetic network $g$ for the oracle based black-box attack from~\cite{PapernotMcDanielGoodfellowEtAl2017}. The attacker uses as input an oracle ${\cal O}$ which represents black-box access to the target model $f$ which only returns the final class label $F(f(x))$ for a query $x$ (and not the score vector $f(x)$). Initially, the attacker has (part of) the training data set ${\cal X}$, i.e., he knows ${\cal D}=\{ (x, F(f(x))) : x\in {\cal X}_0\}$ for some $\mathcal{X}_0\subseteq \mathcal{X}$. Notice that for a single iteration $N=1$, Algorithm~\ref{alg:papernot_synthetic} therefore reduces to an algorithm which does not need any oracle access to ${\cal O}$; this reduced algorithm is the one used in the pure black-box attack~\cite{CarliniWagner2017,AthalyeCarliniWagner2018,LiuChenLiuEtAl2016}. In this paper we assume the strongest black-box adversary in Algorithm~\ref{alg:papernot_synthetic} with access to the entire training data set ${\cal X}_0={\cal X}$ (notice that this excludes test data for evaluating the attack success rate).

In order to construct a synthetic network the attacker chooses a-priori  a substitute architecture $G$ for which the synthetic model parameters $\theta_g$ need to be trained. The attacker uses known image-label pairs in ${\cal D}$ to train $\theta_g$ using a training method $M$ (e.g., Adam~\cite{kingma2014adam}). In each iteration the known data is doubled using the following data augmentation technique: For each image $x$ in the current data set ${\cal D}$,  black-box access to the target model gives label $l={\cal O}(x)$. The Jacobian of the synthetic network score vector $g$ with respect to its parameters $\theta_g$ is evaluated/computed for image $x$. The signs of the column in the Jacobian matrix that correspond to class label $l$ are multiplied with a (small) constant $\lambda$ -- this constitutes a vector which is added to $x$. This gives one new image for each $x$ and this leads to a doubling of ${\cal D}$. After $N$ iterations the algorithm outputs the trained parameters $\theta_g$ for the final augmented data set ${\cal D}$.

\begin{algorithm}%[H]
  \caption{Construction of synthetic network $g$ in Papernot's oracle based black-box attack}
  \label{alg:papernot_synthetic}
\begin{algorithmic}[1]
  \State {\textbf{Input}:} 
     \State \hspace{1cm} $\mathcal{O}$ represents black-box access to $F(f(\cdot))$ for target model $f$ with output function $F$;
     \State \hspace{1cm}  $\mathcal{X}_0\subseteq \mathcal{X}$, where $\mathcal{X}$ is the training data set of target model $f$;
     \State \hspace{1cm} substitute architecture $G$ 
     \State \hspace{1cm}  training method $\mathrm{M}$;
     \State \hspace{1cm}  constant $\lambda$;
     \State \hspace{1cm}  number $N$ of synthetic training epochs 
    \State {\bfseries Output:} 
      \State  \hspace{1cm}  synthetic model $g$ defined by parameters $\theta_g$ 
      \State \hspace{1cm} ($g$ also has output function $F$ which selects the max confidence score; 
      \State \hspace{1cm} $g$ fits architecture $G$)
      \State
      \State {\bfseries Algorithm:}
       %\State  Define $g$
       \For{$N$ iterations}
          \State $\mathcal{D} \leftarrow \{(x,\mathcal{O}(x)): x \in \mathcal{X}_{t}\}$
          \State  $\theta_g = \mathrm{M}(G,\mathcal{D})$
          \State  $\mathcal{X}_{t+1} \leftarrow \{x+\lambda \cdot \mathrm{sgn}(J_{\theta_g}(x)[\mathcal{O}(x)]): x \in \mathcal{X}_t \} \cup \mathcal{X}_t$
       \EndFor
       \State Output $\theta_g$
\end{algorithmic}
\end{algorithm}

The precise set-up for our experiments is given in Tables \ref{tab:method}, \ref{table:setup_datasets}, and \ref{tab:CNN}. Table \ref{tab:method} details the used training method $M$ in Algorithm \ref{alg:papernot_synthetic}. For the evaluated data sets Fashion-MNIST and CIFAR-10 without data augmentation, we enumerate in Table \ref{table:setup_datasets} the amount $|\mathcal{X}_0|$ of training data together with parameters $\lambda$ and $N$ in Algorithm \ref{alg:papernot_synthetic} ($\lambda=0.1$ and $N=6$ are taken from the oracle based black-box attack paper of \cite{PapernotMcDanielGoodfellowEtAl2017}; notice that a test data set of size 10.000 is standard practice; all remaining data serves training and this is {\em entirely} accessible by the attacker). 

Table \ref{tab:CNN} depicts the architecture $G$ of the CNN network of the synthetic network $g$ for the different data sets; the structure has several layers (not to be confused with 'protection layer' in BUZz which is an image transformation together with a whole CNN in itself). The adversary attempts to attack BUZz and will first learn a synthetic network $g$ with architecture $G$ (used as input in Algorithm \ref{alg:papernot_synthetic} that corresponds to Table \ref{tab:CNN}. Notice that the image transformations are kept secret and for this reason the attacker can at best train a synthetic vanilla network. Of course the attacker does know the set from which the image transformations in BUZz are taken and can potentially try to learn a synthetic CNN for each possible image transformation and do some majority vote (like BUZz) on the outputted labels generated by these CNNs. However, there are exponentially many transformations making such an attack infeasible. For future research we will investigate whether a small sized subset of 'representative' image transformations can be used to generate a synthetic model which can be used to attack BUZz in a more effective way. Nevertheless, we believe that BUZz will remain secure because of the security argument given in Section \ref{sec:methods} where is shown how a single perturbation $\eta$ leads to very different perturbations at each protected layer in BUZz. This leads to 'wide' buffer zones and their mere existence is enough to achieve our security goal -- security is not derived from keeping the image transformations private. Keeping these transformations private just makes it harder for the adversary to construct a more effective attack but the resulting attack is expected to still have small attacker's success rates. We leave this study for future work.

\begin{table}[th!]
\caption{Training parameters used in the experiments}
\label{tab:method}
\begin{center}
 \begin{tabular}{|c c|} 
 \hline
 Training Parameter & Value  \\ [0.5ex] 
 \hline\hline
 Optimization Method & ADAM  \\ 
 
 Learning Rate & 0.0001  \\
 
 Batch Size & 64  \\

 Epochs & 100  \\
 
 Data Augmentation & None  \\ [1ex] 
 \hline
\end{tabular}
\end{center}
\label{tab:exsetup}
\end{table}

\begin{table}[h!]
\caption{Mixed black-box attack parameters}
\centering
\scalebox{0.8}{
\begin{tabular}{|c|c|c|c|c|} 
 \hline
  & $|\mathcal{X}_0|$ & $N$ & $\lambda$  \\ [0.5ex] 
 \hline\hline
CIFAR-10& 50000 & 4 & 0.1 \\ \hline
Fashion-MNIST & 60000 & 4 & 0.1 \\ \hline 
\end{tabular}
}
\label{table:setup_datasets}
\end{table}

%This is the synthetic table 
\begin{table}[th!]
\caption{Architectures of synthetic neural networks $g$ from \cite{carlini2017adversarial}}
\label{tab:CNN}
\begin{center}
 \begin{tabular}{|c c |} 
 \hline
 Layer Type & Fashion-MNIST and CIFAR-10 \\ [0.5ex] 
 \hline\hline
 Convolution + ReLU & 3 $\times$ 3 $\times$ 64 \\ 
 
 Convolution + ReLU & 3 $\times$ 3 $\times$ 64   \\
 
 Max Pooling & 2 $\times$ 2  \\
 
 Convolution + ReLU & 3 $\times$ 3 $\times$ 128  \\
 
 Convolution + ReLU & 3 $\times$ 3 $\times$ 128 \\

 Max Pooling & 2 $\times$ 2  \\
 
 Fully Connected + ReLU & 256  \\
 
 Fully Connected + ReLU & 256  \\
 
 Softmax & 10  \\ [1ex] 
 
 \hline
\end{tabular}
\end{center}
\label{tab:networksetup}
\end{table}

\noindent
{\bf White-box attack on the synthetic network.} We perform the white-box attacks as described in Section~\ref{sec:background} such as FGSM~\cite{goodfellow2014explaining}, BIM~\cite{kurakin2017adversarial}, MIM~\cite{dong2018boosting}, PGD~\cite{madry2018towards}, Carlini$\&$Wagner\cite{carlini2017adversarial} and EAD~\cite{chen2018ead} attacks on synthetic model in the mixed black box attacks. The reader can find the description of the attacks in
Appendix~\ref{app-white}.

When a certain white-box attack is used as a pure black-box attack, then no oracle access is available and comparison $l'={\cal O}(x)$ is replaced by comparison $l'=F(g(x))$, which uses the synthetic network.

The parameters of the white-box attacks used in our paper can be found in the following table \ref{tab:attack_parameters}.  
\begin{table}[th!]
\caption{Attacks' parameters. $i$ - number of iterations, $d$ - decaying factor, $r$ radius of the ball for generating the initial noise, $c$ - constant value of C$\&$W attack, $\epsilon$ - noise magnitude, $\beta$ - constant value of EAD attack. Binary Search = Bi.Sr}
\label{tab:attack_parameters}
\centering
\scalebox{0.9}{
\begin{tabular}{|p{1cm}|p{4.5cm}|p{4.5cm}| }
\hline  
Attacks & Fashion-MNIST              & CIFAR-10 \\ \hline \hline
FGSM    & $\epsilon = 0.15$          & $\epsilon=0.05$  \\ \hline 
BIM     & $i=10,\epsilon=0.015$      & $i=10,\epsilon=0.005$ \\ \hline
PGD     & $i=10,r=0.031,\epsilon=0.015$      & $i=10,r=0.031,\epsilon=0.005$ \\ \hline
MIM     & $i=10,d=1.0,\epsilon=0.015$      & $i=10,d=1.0,\epsilon=0.005$ \\ \hline
C$\&$W  & $i=1000,c=\text{Bi.Sr}$      & $i=1000,c=\text{Bi.Sr}$ \\ \hline
EAD  & $i=1000,c=\text{Bi.Sr},\beta=0.01$      & $i=1000,c=\text{Bi.Sr},\beta=0.01$ \\ \hline
%\hline
\end{tabular}
}
\end{table}

\noindent
{\bf Success rate black-box attack.}
In order to implement the black-box attack we first run Algorithm \ref{alg:papernot_synthetic} which outputs the parameters of a synthetic network $g$. Next, out of the test data (each data set has 10.000 samples in our set-up) we  select the first 1000 
%(this is custom in literature) 
samples $(x,l)$ which the target model $f$ (i.e., BUZz in this paper) correctly classifies. For each of the 1000 samples we run a certain white-box attack to produce 1000 adversarial examples. The attacker's success rate is the fraction of adversarial examples which change $l$ to the desired new randomly selected $l'$ in a targeted attack or any other label $l'\neq \bot$ for an untargeted attack.

\noindent
{\bf Image transformations for BUZz.} In the BUZz, we use image transformations that are composed of a resizing operation $i(x)$ and a linear transformation $c(x)=Ax+b$. An input image $x$ at a protected layer in BUZz is linearly transformed into an image $i(c(x))$ before it enters the corresponding CNN network with ResNet architecture for CIFAR-10 and for Fashion-MNIST. In a network implementation one can think of $i(c(x))$ as an extra layer in the CNN architecture of ResNet itself.

For the resize operations $i(\cdot)$ used in each of the protected layers in BUZz, we choose sizes that are larger than the original dimensions of the image data. We do this to prevent  loss of information in the images that down sizing would create (and this would hurt the clean accuracy of BUZz). In our experiments we use BUZz with 2, 4, and 8 protected layers.  Each protected layer gets its own resize operation $i(\cdot)$. When using 8 protected layers, we use image resizing operations from 32 to 32, 40, 48, 64, 72, 80, 96, 104. Each protected layer will be differentiated from each other protected layer due to the difference in how much resizing each layer implements. This will lead to less transferability between the protected layers and as a result we expect to see a wider buffer zone which diminishes the attacker's success rate.
When using 4 protected layers, we use a copy of the 4 protected layers from BUZz with 8 networks that correspond to the  image resizing operations from 32 to 32, 48, 72, 96. 
When using 2 protected layers, we use a copy of the 2 protected layers from BUZz with 8 networks that correspond to the  image resizing operations from 32 to 32 and 104. In our implementation we use  resizing operation from github  \url{https://github.com/cihangxie/NIPS2017_adv_challenge_defense}~\cite{XieWangZhangEtAl2017}. 

For each protected layer, the linear transformation $c(x)=Ax+b$ is randomly chosen from some statistical distribution (the distribution is public knowledge and therefore known by the adversary). Design of the statistical distribution depends on 
the complexity of the considered data set (in our case we experiment with Fashion-MNIST and CIFAR-10).
Transformation $c(x)$ takes an image of size $n_1\times n_2 \times_3$ as input and considers this as a vector of length $k=n_1n_2n_3$. Here, $n_1$ and $n_2$ denote the horizontal and vertical width in pixels of image $x$; $n_3=3$ means that each pixel has a red, blue, and green values; $n_3=1$ means that each pixel only has one black/white value. CIFAR-10 has $32\times 32\times 3$ images and Fashion-MNIST has $28\times 28\times 1$ images. All the values in vector $x$ are converted from integers $[0..255]$ to the range $[-0.5,+0.5]$ of real numbers. Notice that the entries of $c(x)$ may have their values outside of this range.

In our implementation we do not consider $x$ to be in vector representation; we think of $x$ as $n_3$ times a $n_1\times n_2$ matrix. For example, $x=(X_1,X_2,X_3)$ for $n_3=1$. We restrict $c(x)=Ax+b$ to linear operations
$$c(X_1,X_2,X_3) = (X_1A_1+b_1, X_2A_2+b_2, X_3A_3+ b_3),$$
where $A_i$ are $n_2\times n_2$ matrices and $b_i$ are $n_1\times n_2$ matrices.

For CIFAR-10 we take matrices $A_i$ to be identity matrices (this also makes $A$ the identity matrix in the vector representation of $c(x)$) and we use the same matrix $b$ for each of the matrices $b_i$, i.e.,
$$b'=b_1=b_2=b_3.$$
This means that we use the same random offset in the red, blue, and green values of a pixel. 
The reason for making this design decision is because for CIFAR-10 we found that fully random $A$ creates large drops in clean accuracy, even when the network is trained to learn such distortions. As a result, for data sets with high spatial complexity like CIFAR-10, we do not select $A$ randomly. We choose $A$ to be the identity matrix. Likewise for $b'$ we only randomly generate $35\%$ of the matrix values and leave the rest as $0$. For the randomly generated values, we choose them from a uniform distribution from $-0.5$ to $0.5$. 

For datasets with less spatial complexity like Fashion-MNIST, we equate matrices $A'=A_1=A_2=A_3$ and $b'=b_1=b_2=b_3$ and select $A'$ and $b'$ as  random matrices: The values of $A'$  and $b'$ are selected from a Gaussian distribution with $\mu=0$ and $\sigma=0.1$.

\section{Experimental results}
\label{app:exp}

%\textbf{Buffer Zone Graphs.} 
\subsection{Buffer zone graphs} \label{bufferzones}

In Figure~\ref{fig:fish} we show buffer zone graphs for various defenses. These graphs are based on the the decision region graphs originally presented in~\cite{LiuChenLiuEtAl2016}. In our graphs, each point on the 2D grid corresponds to the class label of an image $I'$. Green represents that $I'$ has been classified correctly, while red and blue regions represent incorrect class labels. Gray represents that the null (adversarial) class label has been assigned. The image $I'$ is generated from the original image $I$ using the following equation: $I'=I+x \cdot g+y \cdot r$. Here $g$ represents the gradient of the loss function with respect to $I$. In the equation, $r$ represents a normalized random matrix that is orthogonal to $I$ (note $g$ is also normalized). The other variables, $x$ and $y$ represent the magnitude of each matrix which is determined based on the coordinates in the 2D graph.  

In essence the graph can be interpreted in the following sense: At the origin $I'$ is equal to $I$. The origin is the original image without adversarial perturbations or random noise added. As we move along the  x-axis in the positive direction, the magnitude of the gradient matrix $x$ increases. Moving positively along only the x-axis is equivalent to the FGSM attack, where the image is modified by adding the gradient of the loss function (with respect to the input). If we move along the y-axis only, the magnitude of the random noise matrix $y$ increases. This is equivalent to adding random noise to the image. Moving along the positive x-axis and any direction in the y-axis means we are adding an adversarial perturbation and a random noise to the original image $I$. The further from the origin, the greater the magnitude of $x$ and $y$ and hence the larger the distortion that is applied to create $I'$.

In the case where a defense uses multiple networks, each network $j$ will have a different gradient matrix $g_{j}$. To compensate for this, we  average the individual $g_{j}$ matrices together before normalizing to get $g$. It is important to note that while the graphs shown in Figure~\ref{fig:fish} give experimental proof of the concept of buffer zones, they cannot be used to attack BUZz defenses in practice. When creating the graphs, we have knowledge of the individual gradient matrices $g_{j}$ for each individual network $j$. This information is not available or obtainable by an adversary in a black-box setting, to the best of our knowledge.

\subsection {Attacks}
The hierarchy of black-box attacks we experiment with can be divided into two different types. The first type is generation of the synthetic model using training data and synthetic data labeled by the oracle~\cite{PapernotMcDanielGoodfellowEtAl2017} which we call the mixed black-box attack. The parameters for this attack are given in Table~\ref{table:setup_datasets}. The second type of black-box attack we consider is the pure black-box attack. In this attack the generation of the synthetic model is accomplished using the original training dataset and original training labels. 
The parameters for this attack are given in Table~\ref{table:setup_datasets}. 
In both black-box attacks after the synthetic model is generated, we can run any white-box attack on the synthetic model to create adversarial samples to try and fool the defense. The white-box attacks we consider are FGSM~\cite{goodfellow2014explaining}, BIM~\cite{kurakin2017adversarial}, MIM~\cite{dong2018boosting}, PGD~\cite{madry2018towards}, Carlini$\&$Wagner\cite{carlini2017adversarial} and EAD~\cite{chen2018ead}. The parameters for these attacks are given in Table~\ref{tab:attack_parameters}.

%\begin{table}[th!]
%\caption{Attacks' parameters. $i$ - number of iterations, $d$ - decaying factor, $r$ radius of the ball for generating the initial noise, $c$ - constant value of C$\&$W attack, $\epsilon$ - noise magnitude, $\beta$ - constant value of EAD attack. Binary Search = Bi.Sr}
%\centering
%\scalebox{0.72}{
%\begin{tabular}{|p{1cm}|p{4.5cm}|p{4.5cm}| }
%\hline  
%Attacks & Fashion-MNIST              & CIFAR-10 \\ \hline \hline
%FGSM    & $\epsilon = 0.15$          & $\epsilon=0.05$  \\ \hline 
%BIM     & $i=10,\epsilon=0.015$      & $i=10,\epsilon=0.005$ \\ \hline
%PGD     & $i=10,r=0.031,\epsilon=0.015$      & $i=10,r=0.031,\epsilon=0.005$ \\ \hline
%MIM     & $i=10,d=1.0,\epsilon=0.015$      & $i=10,d=1.0,\epsilon=0.005$ \\ \hline
%C$\&$W  & $i=1000,c=\text{Bi.Sr}$      & $i=1000,c=\text{Bi.Sr}$ \\ \hline
%EAD  & $i=1000,c=\text{Bi.Sr},\beta=0.01$      & $i=1000,c=\text{Bi.Sr},\beta=0.01$ \\ \hline
%\hline
%\end{tabular}
%}
%\label{tab:attack_parameters}
%\end{table}
%%table* for 2 columns

% ==================== (MOVE - END) =======================

\subsection{Experimental analysis}
\label{expana}

% ==================== (MOVE - START) =======================

In Table~\ref{table:pureBB_Fashion_CIFAR} we show the
defense success rate $1-\alpha$
for the pure black-box attack on CIFAR-10 and Fashion-MNIST for various BUZz defenses. Here we do not test every defense as these experimental results are only used to demonstrate two points. The first point is that the pure black-box attack is always equal to or weaker than the mixed black-box attack if you compare these table results to those in    Table~\ref{table:mixBB_Fashion_all_BUZz} and Table~\ref{tab:all_attack_all_defenses_CIFAR10}. The second important point is that the BUZz defense is able to mitigate this type of attack. i.e. the defense accuracy of BUZz-8 is $88\%$ or greater for all attacks.

\begin{figure}%[!h]
\centering
\includegraphics[scale=0.25]{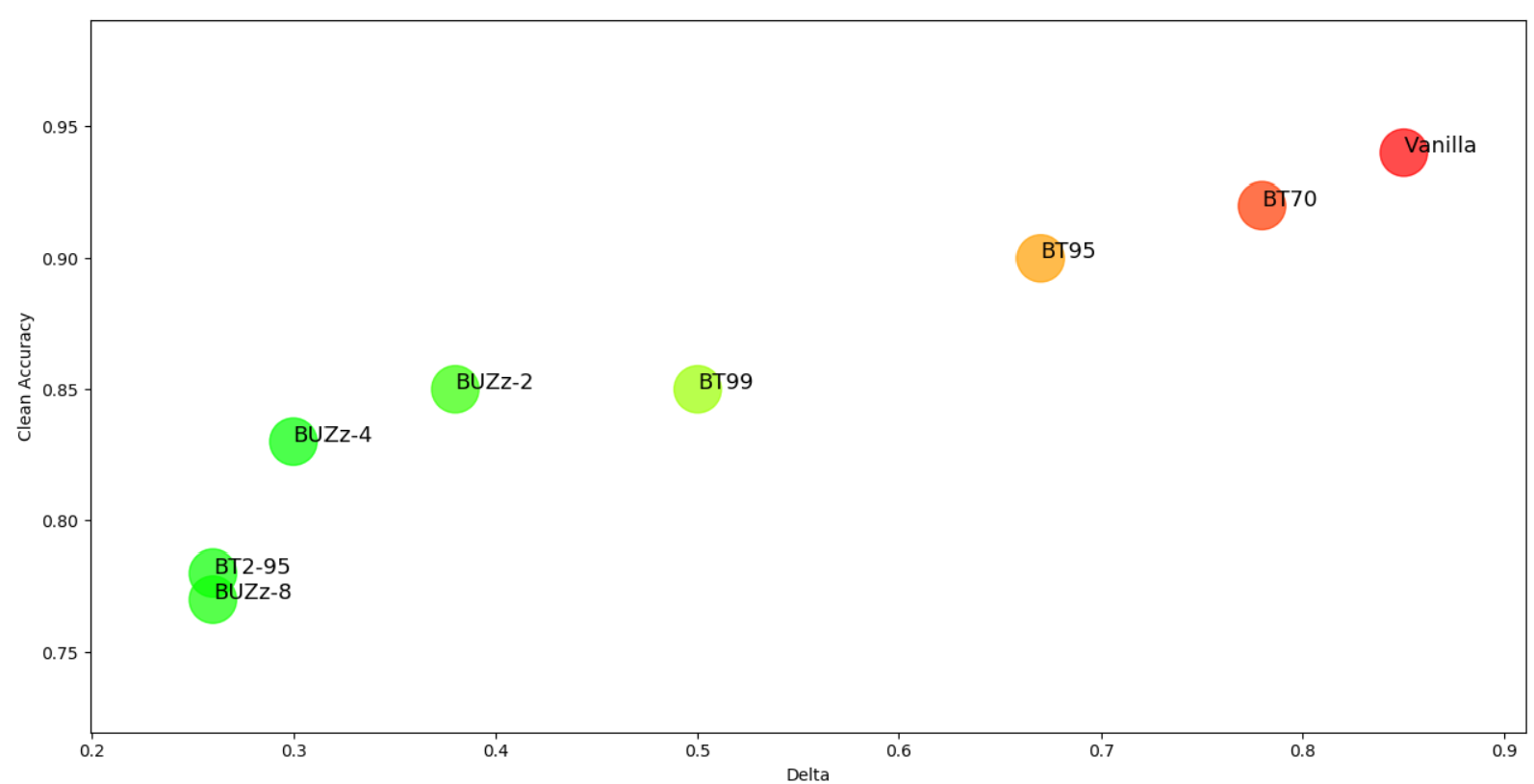}
\caption[]{Fashion-MNIST defenses with clean accuracy $p_d$ (Clean Accuracy) and $\delta$ value (Delta) for untargeted attacks.}
 \label{fig:key_idea3}
%\vspace{2mm}
\noindent
\end{figure}

In Table~\ref{table:mixBB_Fashion_all_BUZz} we show the defense success rate $1-\alpha$ for the mixed black-box attack on Fashion-MNIST for various BUZz defense configurations and a vanilla (no defense) network. In this table we also show the clean accuracy and $\delta$ metric (the best $\delta$ is obtained by the BUZz-8 defense). We can interpret Table~\ref{table:mixBB_Fashion_all_BUZz} by Figure \ref{fig:key_idea3}.

\noindent 

In Table~\ref{tab:all_attack_all_defenses_CIFAR10}, for CIFAR-10 we show the clean accuracy $p_d$ and defense success rate $1-\alpha$ for various attacks. We also compute $\delta$ and the drop in accuracy $\gamma$ for some defenses such as BUZz 2, 4, 8, BUZz using threshold cutoff 0.7, 0.95 and 0.99, the combination BT2-70, and plain classifier (Vanilla), Xie's: Randomized 1-Net~\cite{XieWangZhangEtAl2017}, Liu's: Mixed Arch 2-Net~\cite{LiuChenLiuEtAl2016}, Cao's: randomized smooth technique~\cite{CaoGong2017,cohen2019certified},  Guo's: this is BUZz with a single network~\cite{GuoRanaCisseEtAl2017}, Tramer's: adversarial training~\cite{TramerKurakinPapernotEtAl2017}, MD2 and MD4~\cite{srisakaokul2018muldef}. In Table~\ref{tab:all_attack_all_defenses_CIFAR10} the $\delta$ value is computed based on the attack success rate of the most successful attack – the targeted and untargeted, mixed black box attack with MIM. For example, to compute the $\delta$ for the vanilla case, we see that the untargeted MIM attack is the most efficient one with attack success rate $\alpha$ of $0.72$ (=1-0.28), $p=0.93$ and $p_d=p=0.93$, thus $\delta = 0.93 - 0.93\times 0.28 =0.66$ and $\gamma = p - p_d = 0$. For BUZz2, since $p=0.93$, $p_d=0.85$ and $1-\alpha= 0.68$, $\delta = 0.93 - 0.85\times 0.68 = 0.36$ and $\gamma = 0.93 - 0.85 = 0.07$. 

 %We reported our experimental results for CIFAR-10 in Section \ref{sec:experiment} for 5 runs (see Table~\ref{tab:all_attack_all_defenses_CIFAR10}. 
For our BUZz defense, one run is implemented by first choosing matrices $A'$ and $B'$ from the distribution corresponding to the considered data set for each protected network. Next the attacker's success rate and clean accuracy of the defense are simulated. For each next run, matrices $A'$ and $B'$ are again chosen anew. 
% As shown in Figure~\ref{table:mixBB_all_defenese_CIFAR} for CIFAR-10, the average result for 5 runs is not much different from that of 1 run. 
In Table~\ref{tab:all_attack_all_defenses_CIFAR10}, we report the result of the best case of the attack (which is the worst case for the defender) among 5 runs. Indeed, all the attack results are not much different, i.e., they are only different in 1$\%$. This shows that BUZz is not sensitive to the choice of $A'$ and $B'$ (worst and best cases are close to one another).

\begin{figure}%[!h]
\centering
\includegraphics[scale=0.25]{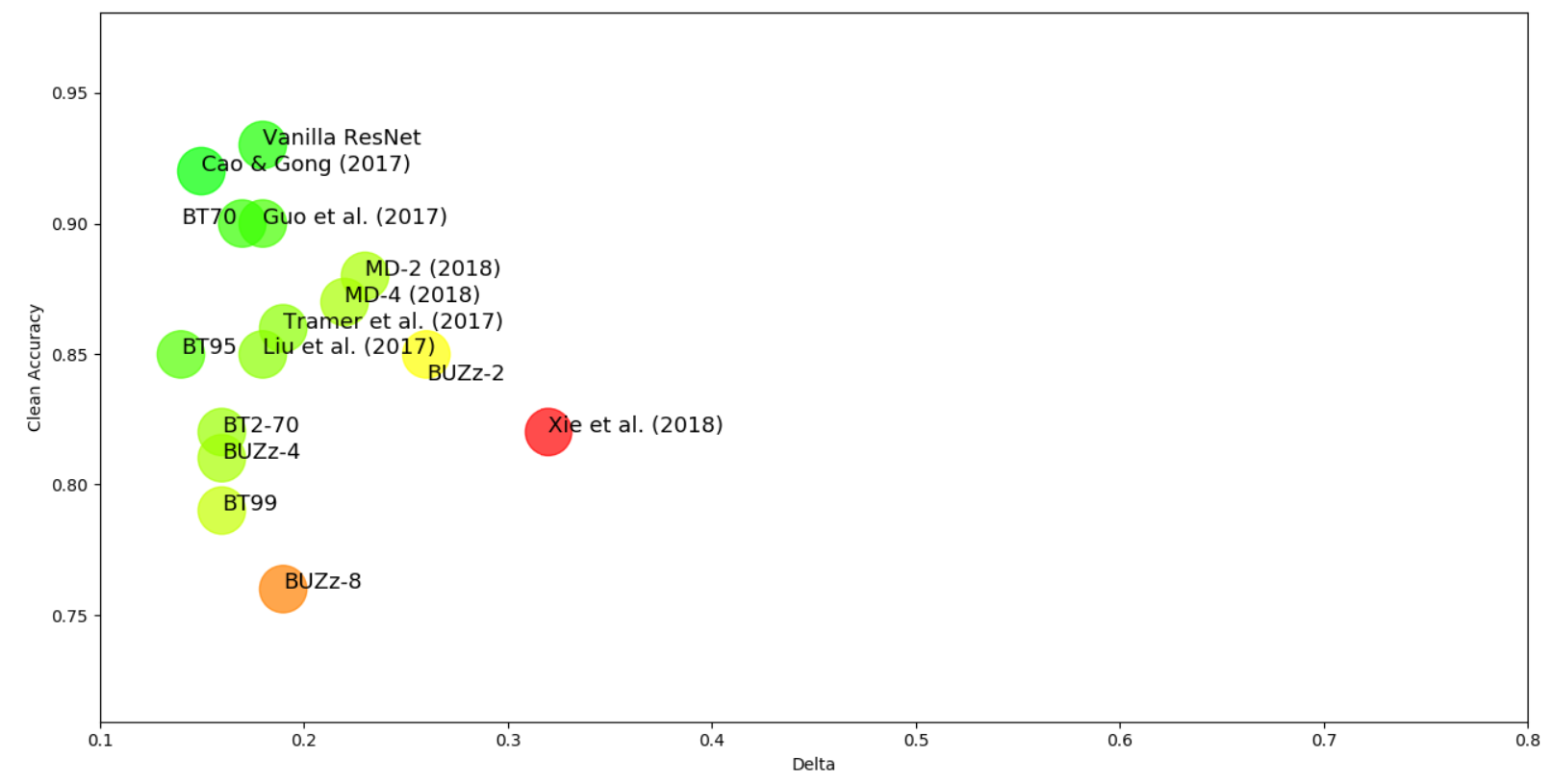}
\caption[]{CIFAR-10 defenses with clean accuracy $p_d$ (Clean Accuracy) and $\delta$ value (Delta) for targeted attacks.}
 \label{fig:key_idea2}
%\vspace{2mm}
\noindent
\end{figure}

We can interpret Table~\ref{tab:all_attack_all_defenses_CIFAR10} by Figures \ref{fig:key_idea} and \ref{fig:key_idea2} for untargeted and targeted attacks. We see that for the broader class of untargeted attacks BUZz offers a significantly better defense when compared to our selection of  most prominent defenses in literature. On the other hand we see that for targeted attacks having no defense at all (the vanilla scheme) is almost as good as the best defenses given by Cao \& Cong and BUZz strategy BT70. 

For targeted attacks we have two remarks: First, in a targeted attack adversarial noise must already bridge other non-targeted labeled regions in a vanilla scheme and as such it already behaves like a BUZz scheme. This explains why implementing no attack at all for targeted attacks behaves like our BUZz scheme. Second, among the vanilla scheme and the other best defenses the $\delta$ value and clean accuracy $p_d$ are  approximately the same. Among these defenses, we can now choose which attack accuracy $\alpha$ is preferred. In this case the vanilla scheme has  a defense success rate $1-\alpha=0.81$, while BUZz's BT70 and Cao \& Cong have $1-\alpha = 0.84$ and $0.85$ close to one another.

The results for the mixed black-box attacks on Fashion-MNIST and CIFAR-10 are presented in Table~\ref{table:mixBB_Fashion_all_BUZz} and 
Table~\ref{tab:all_attack_all_defenses_CIFAR10}.
%Table~\ref{table:pureBB_Fashion_CIFAR}. 
Overall we can see that the different BUZz configurations offer better defense accuracy than the existing literature defenses. Specifically, if we consider BUZz-8, the attack success rate drops to 7\% for the strongest black-box adversary on CIFAR-10 and to 11\% for Fashion-MNIST. However, this security comes with a drop in clean accuracy to 76\% for CIFAR-10 and 77\% for Fashion-MNIST. Alternatively, other BUZz defenses can be chosen if a higher clean accuracy is desired. For example, using BUZz with a threshold cutoff of 0.7 for CIFAR-10 gives a defense success rate of 52\% and a clean accuracy of 90\%.

\subsection{Discussion} 

From the experiments on Fashion-MNIST and CIFAR-10, we see the success of untargeted mixed black-box attacks on vanilla nets is significantly higher than that of the targeted attacks. 

% Moreover,  the success of targeted attacks on the vanilla network of CIFAR-100 is close to $1\%$ which is significantly poorer than that of CIFAR-10 or FashionMNIST (i.e., $16\%-19\%$). We believe the main reason is that the clean accuracy of the vanilla network of CIFAR-100 is significantly lower than that of CIFAR-10 or FashionMNIST, i.e., $63\%, 88\%$ and $93.5\%$, respectively. From this result, we also believe that the resistance of a network or classifier against adversarial machine learning strongly depends on the clean accuracy of the network itself; the lower the clean accuracy, the lower the attacker's success rate as well (remember that the attacker's success rate is only measured over images that are accurately predicted by the network).   

From the view of the designer, she/he should pay attention to the highest threat or the most powerful attack, in this case untargeted mixed black-box attacks. In other words, we should use the $\delta$ of untargeted mixed black-box attacks as a true measurement to evaluate the resistance of a given defense. Equipped with this argument, we can now see that BUZz has very good resistance to adversarial examples %in general case
at the cost of clean accuracy. 
As we argued in Table~\ref{table:success_rate_bb},
%~\ref{table:delta}, 
current-state-of-the-art defenses 
%discussed in this paper has 
generally have very large $\delta$ which is due to poor resistance against adversarial examples --
%is significantly poor but
this is because 
their clean accuracy is nearly equal to that of the vanilla network (these defenses do not want to give up some of the clean accuracy). A good lesson we can learn from Table~\ref{table:success_rate_bb} is that we have to sacrifice something (here, clean accuracy) to gain security.
%and it seems the chosen one should be 'prediction accuracy'. 
If the defense does not have `buffer zones', then the adversary always wins the game in that it is possible to produce with significant probability adversarial examples with small noise to bypass the defense. Including `buffer zones' means the designer has to give up some clean accuracy.   

For Fashion-MNIST, we may want to use 8-network BUZz because it has the best trade-off between defense rate (one minus the attacker's succes rate) and the defense accuracy (the clean accuracy of the defense). Similarly, for CIFAR-10 we suggest 4-network BUZz.

%=====================

In this paper the main strength of BUZz is two-fold. First, we provide the highest defense accuracy  (as compared to other defenses in the literature) with the defense configuration BUZz-8. Second, we offer the user a spectrum of choices. We do not claim any one BUZz defense is superior but let the user choose the defense based on their own values (whether they prioritize defense accuracy  or clean accuracy $p_d$).  

%TO DO: Explain that delta metric obfuscates the exchange of $\gamma$ and $\alpha$: increasing $\gamma$ and lowering $\alpha$ can lead to the same $\delta$. It depends on how you want to exchange these.

\noindent

\begin{table}[th!]
  \centering
\caption{Targeted (Tar.) and Untargeted (Untar.) pure black-box attacks on BUZz 2, 4, 8 defenses for Fashion-MNIST and CIFAR-10. Clean prediction accuracy $p_d$ - Cl.Ac $p_d$, Drop in clean prediction accuracy $\gamma$. }
\scalebox{0.6}{
\begin{tabular}{ |p{1.7cm}|p{1cm}|p{1cm}|p{1cm}|p{1cm}|p{1cm}|p{1cm}|p{1cm}|p{1cm}|   }
\hline
&\multicolumn{4}{c}{Fashion-MNIST} &\multicolumn{4}{|c|}{CIFAR-10}  \\
%\hline
Attack/Type & Vanilla & BUZz2 & BUZz4 & BUZz8 & Vanilla & BUZz2 & BUZz4 & BUZz8  \\
\hline \hline
FGSM/Tar.  & 0.87 & 0.95 & 0.97 & 0.98 & 0.90 & 0.94 & 0.97 & 0.99  \\
FGSM/Untar.  & 0.43 & 0.75 & 0.82 & 0.88  & 0.44 & 0.75 & 0.87 & 0.94  \\

\hline
BIM/Tar. & 0.89 & 0.99 & 0.99 & 0.99  & 0.92 & 0.96 & 0.99 & 0.99  \\

BIM/Untar. & 0.36 & 0.82 & 0.88 & 0.93 & 0.45 & 0.73 & 0.86 & 0.93  \\

\hline
PGD/Tar.  & 0.88 & 0.99 & 0.99 & 0.99  & 0.92 & 0.97 & 0.99 & 0.99 \\

PGD/Untar.  & 0.37 & 0.83 & 0.87 & 0.92  & 0.45 & 0.75 & 0.85 & 0.93  \\

\hline
MIM/Tar.  & 0.82 & 0.97 & 0.99 & 0.99 & 0.85 & 0.93 & 0.97 & 0.99  \\

MIM/Untar.  & 0.35 & 0.78 & 0.85 & 0.90  & 0.38 & 0.70 & 0.84 & 0.90  \\

\hline
C$\&$W/Tar.& 0.99 & 0.99 & 1.0 & 1.0 & 0.98 & 0.99 & 0.99 & 1.0  \\
C$\&$W/Untar.& 0.91 & 0.95 & 0.97 & 0.99  & 0.92 & 0.96 & 0.98 & 1.0  \\

\hline
EAD/Tar. & 0.99 & 0.99 & 1.0 & 1.0 & 0.98 & 0.99  & 0.99 & 1.0  \\

EAD/Untar. & 0.90 & 0.95 & 0.97 & 0.99 & 0.92 & 0.96 & 0.98 & 1.0  \\
\hline

Cl.Ac $p_d$  & 0.94 & 0.85 & 0.83 & 0.77 & 0.93 & 0.85 & 0.81 & 0.76 \\

\hline 

$\delta$ &  0.62 & 0.31 & 0.28 & 0.27 &  0.58 & 0.34 & 0.25 & 0.25 \\

\hline 

Dp.Ac $\gamma$ &   0.0 & 0.09 & 0.11 & 0.17 &   0.0 & 0.09 & 0.13 & 0.19 \\

\hline 

\end{tabular}
}
\label{table:pureBB_Fashion_CIFAR}
\end{table}

\begin{table}[th!]
\centering
\caption{Targeted (Tar.) and Untargeted (Untar.) mixed black-box attacks on BUZz 2, 4, 8, BUZz has only one single classifier with threshold 0.70 (BT70), 0.95 (BT95) and 0.99 (BT99), BUZz2 having one classifier with threshold 0.95 (BT2-95)  defenses for Fashion-MNIST. Clean prediction accuracy $p_d$ - Cl.Ac $p_d$, Drop in clean prediction accuracy~$\gamma$.}
\scalebox{0.6}{
\begin{tabular}{ |p{1.7cm}|p{1cm}|p{1cm}|p{1cm}|p{1cm}|p{1cm}|p{1cm}|p{1cm}|p{1.2cm}|   }
\hline
%&\multicolumn{5}{c}{\textbf{Fashion MNIST}} &\multicolumn{5}{|c|}{\textbf{CIFAR-10}}  \\
%\hline
Attack/Type & Vanilla & BUZz2 & BUZz4 & BUZz8 & BT70 & BT95 & BT99 & BT2-95 \\
\hline \hline
FGSM/Tar.  & 0.67 & 0.93 & 0.97 & 0.98 & 0.76  & 0.85 & 0.94  & 0.97 \\
FGSM/Untar.  & 0.17 & 0.67 & 0.80 & 0.89 & 0.35 &  0.57 & 0.72 & 0.85   \\

\hline
BIM/Tar. & 0.50 & 0.96 & 0.99 & 1.00 & 0.60 & 0.77 & 0.89 & 0.99   \\

BIM/Untar.& 0.11 & 0.74 & 0.85 & 0.94 & 0.17 & 0.30 & 0.50 & 0.89  \\

\hline
PGD/Tar.  & 0.50 & 0.96 & 0.98 & 0.99 & 0.59 & 0.79 & 0.89 & 0.99 \\

PGD/Untar.  & 0.10 & 0.74 & 0.85 & 0.95 &  0.18 & 0.30 & 0.50 & 0.89  \\

\hline
MIM/Tar. & 0.43 & 0.95 & 0.97 & 0.99 &  0.52 & 0.72 & 0.86 & 0.97 \\

MIM/Untar.  & 0.10 & 0.71 & 0.82 & 0.92 & 0.18 & 0.34 & 0.52 & 0.86 \\

\hline
C$\&$W/Tar. & 0.97 & 1.0 & 1.0 & 1.0 & 0.99 & 1.0 & 1.0 & 1.0 \\
C$\&$W/Untar.  & 0.95 & 1.0 & 1.0 & 1.0 & 0.97 & 1.0 & 1.0 & 1.0 \\

\hline
EAD/Tar.  & 0.98 & 1.0 & 1.0 & 1.0 & 0.99 & 1.0 & 1.0  & 1.0 \\

EAD/Untar.  & 0.93 & 1.0 & 1.0 & 1.0 & 0.95 & 1.0 & 1.0  & 1.0 \\
\hline

Cl.Ac $p_d$  & 0.94 & 0.85 & 0.83 & 0.77 &  0.92 & 0.90 & 0.85 & 0.78 \\

\hline

$\delta$ untar. & 0.85 & 0.38 & 0.30 & 0.26 &  0.78 & 0.67 & 0.5 &  0.26 \\

\hline 

Dp.Ac $\gamma$  & 0.0 & 0.09 & 0.11 & 0.17 &  0.02 & 0.04 & 0.09 & 0.15 \\

\hline 

\end{tabular}
}

\label{table:mixBB_Fashion_all_BUZz}
\end{table}

\begin{table}[th!]
\caption{Targeted (Tar.) and Untargeted (Un.) mixed black-box attacks on common defenses for CIFAR-10. BUZz 2, 4, 8, and BUZz has only one single classifier with threshold 0.7 (BT70), 0.95 (BT95) and 0.99 (BT99), BUZz2 having one classifier with threshold 0.70 (BT2-70), Vanilla: the plain target model, Xie's: Randomized 1-Net~\cite{XieWangZhangEtAl2017}, Liu's: Mixed Arch 2-Net~\cite{LiuChenLiuEtAl2016}, Cao's: randomized smooth function~\cite{CaoGong2017,cohen2019certified}, Guo's: this is BUZz with a single network~\cite{GuoRanaCisseEtAl2017}, Tramer's: adversarial training~\cite{TramerKurakinPapernotEtAl2017}, MD2 and MD4~\cite{srisakaokul2018muldef}. Clean prediction accuracy $p_d$ - Cl.Ac $p_d$, Drop in clean prediction accuracy $\gamma$.}
\centering
\scalebox{0.7}{
\begin{tabular}{|p{1.2cm}|p{0.7cm}|p{0.7cm}|p{0.7cm}|p{0.7cm}|p{0.7cm}|p{0.7cm}|p{0.7cm}|p{0.7cm}|p{0.7cm}|  }
\hline  
Defense & Tar. FGSM & Tar. BIM  & Tar. PGD  & Tar.  MIM  & Tar. C$\&$W& Tar. EAD  & Cl.Ac $p_d$ & $\delta$ & Dp.Ac $\gamma$ \\ \hline \hline
Vanilla & 0.88 & 0.86 & 0.88 & 0.81 & 0.99 & 0.99 & 0.93 &  0.18 & 0.0 \\ \hline
BUZz2   & 0.94 & 0.95 & 0.94 & 0.90 & 1.0  & 1.0  & 0.85 &  0.16 & 0.07 \\ \hline
BUZz4   & 0.97 & 0.97 & 0.97 & 0.95 & 1.0  & 1.0  & 0.81 &  0.16 & 0.12 \\ \hline
BUZz8   & 0.99 & 0.99 & 0.99 & 0.97 & 1.0  & 1.0  & 0.76 &  0.19 & 0.17  \\ \hline
BT70    & 0.91 & 0.89 & 0.89 & 0.84 & 0.99 & 0.99 & 0.90 &  0.17 & 0.03 \\ \hline 
BT95    & 0.96 & 0.95 & 0.96 & 0.93 & 1.0  & 1.0  & 0.85 &  0.14 & 0.07 \\ \hline 
BT99    & 0.98 & 0.98 & 0.98 & 0.97 & 1.0  & 1.0  & 0.79 &  0.16 & 0.14 \\ \hline 
BT2-70 & 0.96 & 0.97 & 0.97 & 0.94 & 1.0  & 1.0  & 0.82 &  0.16 & 0.11 \\ \hline 
Xie's   & 0.86 & 0.83 & 0.83 & 0.75 & 0.98 & 0.97 & 0.82 &  0.32 & 0.11 \\ \hline 
Liu's   & 0.94 & 0.94 & 0.94 & 0.88 & 1.0  & 1.0  & 0.85 & 0.18 & 0.07 \\ \hline 
Cao's   & 0.89 & 0.90 & 0.89 & 0.85 & 0.99 & 0.99 & 0.92 &  0.15 & 0.01  \\ \hline 
Guo's   & 0.89 & 0.90 & 0.90 & 0.83 & 0.99 & 0.99 & 0.90 & 0.18 & 0.03 \\ \hline 
Tramer's& 0.90 & 0.93 & 0.86 & 0.93 & 0.99 & 0.99 & 0.86 & 0.19 & 0.07 \\ \hline 
MD2     & 0.89 & 0.88 & 0.88 & 0.79 & 0.98 & 0.98 & 0.88 & 0.23 & 0.05 \\ \hline 
MD4     & 0.89 & 0.90 & 0.91 & 0.82 & 0.98 & 0.99 & 0.87 & 0.22 & 0.06 \\ \hline \hline
% \end{tabular}
% \begin{tabular}{|p{1.2cm}|p{0.7cm}|p{0.7cm}|p{0.7cm}|p{0.7cm}|p{0.7cm}|p{0.7cm}|p{0.7cm}|p{0.7cm}|p{0.7cm}|  }
% \hline 
Defense & Un. FGSM & Un. BIM  & Un. PGD  & Un. MIM  &Un. C$\&$W& Un. EAD  & Cl.Ac $p_d$ & $\delta$& Dp.Ac $\gamma$ \\ \hline \hline
Vanilla & 0.33 & 0.38 & 0.38 & 0.28 & 0.98 & 0.98 & 0.93 & 0.66 & 0.0 \\ \hline
BUZz2   & 0.75 & 0.76 & 0.76 & 0.68 & 1.0  & 1.0  & 0.85 & 0.36 & 0.07 \\ \hline
BUZz4   & 0.90 & 0.90 & 0.88 & 0.86 & 1.0  & 1.0  & 0.81 & 0.23 & 0.12 \\ \hline
BUZz8   & 0.94 & 0.94 & 0.94 & 0.93 & 1.0  & 1.0  & 0.76 & 0.22 & 0.17 \\ \hline
BT70    & 0.55 & 0.59 & 0.59 & 0.52 & 0.99 & 0.99 & 0.90 & 0.46 & 0.03 \\ \hline 
BT95    & 0.80 & 0.81 & 0.82 & 0.77 & 1.0  & 1.0  & 0.85 & 0.27 & 0.07 \\ \hline 
BT99    & 0.93 & 0.92 & 0.90 & 0.89 & 1.0  & 1.0  & 0.79 & 0.22 & 0.14 \\ \hline 
BT2-70 & 0.86 & 0.86 & 0.86 & 0.80 & 1.0  & 1.0  & 0.82 & 0.27 & 0.11 \\  \hline 
Xie's   & 0.36 & 0.41 & 0.39 & 0.27 & 0.88 & 0.88 & 0.82 & 0.70 & 0.11 \\ \hline 
Liu's   & 0.73 & 0.70 & 0.70 & 0.63 & 0.99 & 0.99 & 0.85 & 0.39 & 0.07 \\ \hline 
Cao's   & 0.42 & 0.48 & 0.48 & 0.39 & 0.98 & 0.98 & 0.92 & 0.57 & 0.01 \\ \hline 
Guo's   & 0.48 & 0.57 & 0.56 & 0.44 & 0.99 & 0.98 & 0.90 & 0.53 & 0.03 \\ \hline 
Tramer's& 0.55 & 0.63 & 0.45 & 0.64 & 0.99 & 0.97 & 0.86 & 0.54 & 0.07 \\ \hline 
MD2     & 0.44 & 0.50 & 0.49 & 0.37 & 0.94 & 0.93 & 0.88 & 0.60 & 0.05 \\ \hline 
MD4     & 0.49 & 0.53 & 0.54 & 0.37 & 0.93 & 0.92 & 0.87 & 0.60 & 0.06 \\ \hline 
%\hline
\end{tabular}
}
\label{tab:all_attack_all_defenses_CIFAR10}
\end{table}

% \begin{figure}[th!]
% \centering
% \includegraphics[scale=0.8]{./Figs/KM/pureFM.pdf}
% \caption[]{Pure Black-box Attack on FashionMNIST, single run.}
%  \label{fig:PureFMNIST}
% \vspace{2mm}
% \noindent
% \end{figure}

% \begin{figure}[th!]
% \centering
% \includegraphics[scale=0.8]{./Figs/KM/pureC10.pdf}
% \caption[]{Pure Black-box Attack on CIFAR-10, single run.}
%  \label{fig:PureC10}
% \vspace{2mm}
% \noindent
% \end{figure}

% \begin{figure}[!h]
% \centering
% \includegraphics[scale=0.8]{./Figs/KM/pureC100.pdf}
% \caption[]{Pure Black-box Attack on CIFAR-100, single run.}
%  \label{fig:PureC100}
% \vspace{2mm}
% \noindent
% \end{figure}

\end{document}